\documentclass[letterpaper,journal]{IEEEtran}
\usepackage{amsmath,amsfonts,amssymb}
\usepackage{bm}
\usepackage{algorithmic}
\usepackage{algorithm}
\usepackage{array}
\usepackage[caption=false,font=footnotesize]{subfig}
\DeclareCaptionFormat{configaxescenter}{%
  \makebox[\linewidth][c]{\hspace*{0.19917092\linewidth}#1#2#3}\par}
\usepackage{textcomp}
\usepackage{stfloats}
\usepackage{url}
\usepackage{verbatim}
\usepackage{graphicx}
\usepackage{xcolor}
\usepackage{tikz}
\usetikzlibrary{shapes.geometric}
\usepackage{cite}
\usepackage[hidelinks]{hyperref}
\definecolor{residualZero}{RGB}{0,114,178}
\definecolor{residualThirty}{RGB}{230,159,0}
\definecolor{residualSixty}{RGB}{213,94,0}
\definecolor{residualPooled}{RGB}{42,42,42}
\newcommand{\residuallegendmark}[3]{%
  \tikz[baseline=-0.5ex,x=1em,y=1em]{%
    \draw[draw=#1,line width=0.9pt,#2] (0,0) -- (2.6,0);
    \node[draw=#1,line width=0.9pt,fill=none,inner sep=0pt,
      minimum size=0.65em,#3] at (1.3,0) {};}}
\definecolor{simDesired}{RGB}{32,33,36}
\definecolor{simNominal}{RGB}{198,40,40}
\definecolor{simSSI}{RGB}{0,114,178}
\definecolor{simOurs}{RGB}{255,176,0}
\definecolor{simForce}{RGB}{251,243,222}
\newcommand{\simlegendline}[2]{%
  \raisebox{0.35ex}{\textcolor{#1}{\rule{1.75em}{#2}}}}
\newcommand{\simlegendpatch}[1]{%
  \raisebox{0.05ex}{\textcolor{#1}{\rule{1.75em}{0.65em}}}}
\definecolor{hardwareCircle}{HTML}{C62828}
\definecolor{hardwareStatic}{HTML}{0072B2}
\definecolor{hardwareDynamic}{HTML}{FFB000}
\definecolor{hardwareDoor}{HTML}{7E2F8E}

\definecolor{residualSigmaZero}{HTML}{7B3294}
\definecolor{residualSigmaInflated}{HTML}{008B8B}
\newcommand{\residualcovlegend}[2]{%
  \tikz[baseline=-0.5ex]{%
    \draw[color=#1,line width=0.9pt,#2]
      (0,0) -- (1.8em,0);
  }%
}
\newtheorem{problem}{\upshape\bfseries Problem}
\newtheorem{proposition}{\upshape\bfseries Proposition}
\newtheorem{theorem}{\upshape\bfseries Theorem}

\newtheorem{lemma}{\upshape\bfseries Lemma}
\newtheorem{assumption}{\upshape\bfseries Assumption}

\newtheorem{definition}{\upshape\bfseries Definition}
\AddToHook{env/problem/begin}{\itshape}
\AddToHook{env/proposition/begin}{\itshape}
\AddToHook{env/theorem/begin}{\itshape}
\AddToHook{env/corollary/begin}{\itshape}
\AddToHook{env/lemma/begin}{\itshape}
\AddToHook{env/assumption/begin}{\itshape}
\AddToHook{env/remark/begin}{\itshape}
\AddToHook{env/definition/begin}{\itshape}
\begin{document}
\title{Online Learning of Heterogeneous Locomotion Dynamics under Reinforcement Learning Policies for Precise Loco-Manipulation}

\author{Tianlin Zhang, Weipeng Xia, Lingwei Zhang, Yiyang Shao, Zhongyu Li,
        and Yun-Hui Liu,~\IEEEmembership{Fellow,~IEEE}%
\thanks{This work is supported in part by the CUHK T Stone Robotics Institute and
in part by the InnoHK initiative of the Innovation and Technology Commission
of the Hong Kong Special Administrative Region Government via the Hong
Kong Centre for Logistics Robotics.}%
\thanks{Corresponding author: Yun-Hui Liu (e-mail: \href{mailto:yhliu@cuhk.edu.hk}{yhliu@cuhk.edu.hk}).}%
\thanks{Tianlin Zhang, Weipeng Xia, Lingwei Zhang, Zhongyu Li, and Yun-Hui Liu are with the Department of Mechanical and Automation Engineering, The Chinese University of Hong Kong, Hong Kong SAR, China (e-mail: \href{mailto:skywoodszcn@gmail.com}{skywoodszcn@gmail.com}).}%
\thanks{Yiyang Shao is with the Hong Kong Centre for Logistics Robotics, Hong Kong SAR, China.}}

\markboth{}%
{Zhang \MakeLowercase{\textit{et al.}}: Online Learning of Heterogeneous Locomotion Dynamics under Policy Control for Precise Loco-Manipulation}


\maketitle

\begin{abstract}
How to reuse a pretrained locomotion policy for precise loco-manipulation is still an open problem. 
We propose an online method to learn the closed-loop dynamics under a frozen locomotion policy. 
The learned model predicts floating-base motion and improves loco-manipulation accuracy. 
However, unobserved latent conditions of the policy introduce heterogeneous responses, 
meaning that the same observed state will lead to different motions. 
Direct learning without accounting for this heterogeneity introduces bias and fluctuations 
into dynamics estimates, degrading floating-base prediction and end-effector tracking.
To address this problem, a contaminated Gaussian model is proposed to approximate 
the effects of latent conditions. The model uses two Gaussian components to approximate predictable 
and unpredictable responses under the same observed state.
Then, the contaminated Gaussian model is combined with a Gaussian process to construct 
an online learner using random Fourier features (RFFs). 
The learner estimates each response’s probability of belonging to the predictable component 
and uses it to weight online updates, capturing predictable responses 
while reducing bias and fluctuations caused by heterogeneous observations.
With consistent and unbiased dynamics estimates, model predictive control (MPC) predicts 
floating-base motion accurately and enables precise loco-manipulation. 
A finite-sample error bound is established to ensure uniform convergence in online learning. 
Simulation and hardware experiments demonstrate consistent end-effector 
tracking across different manipulator configurations and tasks, with errors of 0.01 m.
\end{abstract}


\section{Introduction}
\begin{figure}[!t]
\centering
\includegraphics[width=\columnwidth]{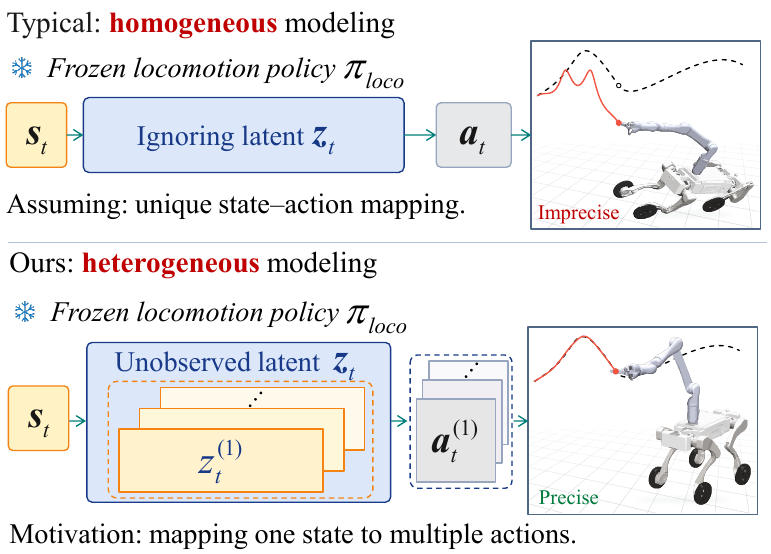}
\caption{Motivation for online learning heterogeneous closed-loop dynamics to achieve
precise loco-manipulation with a frozen locomotion policy.
Unobserved latent conditions $\bm{z}_t$ will produce different actions
$\bm{a}_t$ and closed-loop responses for the same observed state
$\bm{s}_t$, challenging the unique-response assumption of homogeneous
modeling.
Our method accounts for this heterogeneity during online dynamics
learning, enabling precise loco-manipulation without retraining
the locomotion policy.
}
\label{fig:paper_cover}
\end{figure}

\IEEEPARstart{T}{he} heterogeneity of closed-loop dynamics under frozen
reinforcement learning (RL) locomotion policies has been largely overlooked,
degrading tracking accuracy when these policies are reused for loco-manipulation.
Although policy parameters remain frozen, unobserved latent states like
environmental conditions will produce different closed-loop responses under
the same observed state, as shown in Fig.~\ref{fig:paper_cover}.
Ignoring this heterogeneity leads to inaccurate locomotion prediction and
feedforward compensation.
The resulting errors propagate through whole-body kinematics to the end
effector, limiting manipulation precision. This motivates the central
question of this work: \emph{How can the heterogeneity of closed-loop dynamics
under a frozen RL locomotion policy be accounted for to achieve precise
loco-manipulation?}

\begin{figure*}[!t]
\centering
\includegraphics[width=\textwidth]{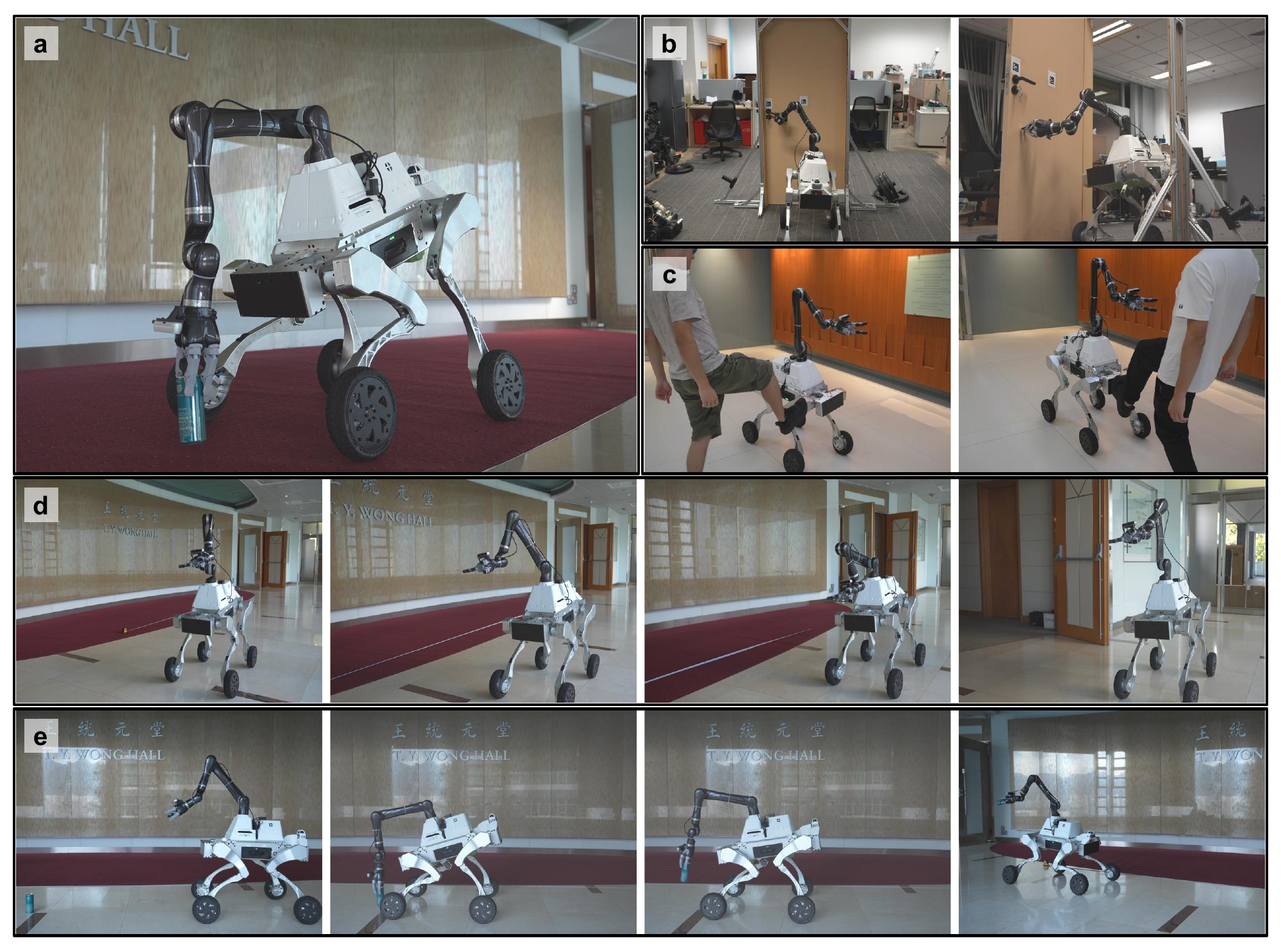}
\caption{Hardware demonstrations of the proposed method, which learns locomotion
dynamics online to enable reuse of the frozen locomotion policy
and generalization across diverse loco-manipulation tasks without
retraining: (a) object grasping, (b) door opening,
(c) disturbance recovery, (d) end-effector trajectory tracking,
and (e) manipulation on the move.
}
\label{fig:hardware_show}
\end{figure*}

Legged loco-manipulation combines the mobility of legged robots with the
dexterity of manipulators, enabling whole-body interaction over workspaces
that cannot be reached from a fixed support \cite{zhang2023dynamic} and
thereby attracting substantial research attention.
Recent advances in reinforcement learning (RL) have substantially advanced
legged locomotion, enabling robust controllers to operate across diverse
conditions \cite{frey2026advances,rudin2022learning}.
This progress, however, has not yet translated into equally mature
loco-manipulation.
Existing approaches still face challenges in achieving precise
loco-manipulation with sample efficiency while accommodating conflicting
task objectives \cite{ha2025learning,gu2026humanoid}.
Sample-based policy optimization explores a high-dimensional state--action
space, whereas exact end-effector pose constraints, under regular kinematic
conditions, restrict feasible configurations to a lower-dimensional manifold
of zero measure in the full configuration space \cite{kingston2019exploring}.
With finite tracking tolerances, admissible configurations occupy a
neighborhood that narrows as the required precision increases, making
successful samples difficult to obtain through unguided exploration.
Conflicts between locomotion and manipulation further complicate policy
optimization.
For example, efficient locomotion generally favors a nearly horizontal
floating-base, whereas end-effector tracking may require base tilting and
fine posture regulation.
Increasing the emphasis on one objective will therefore degrade the
other~\cite{vijayan2025multi}.
To mitigate this conflict, advantage mixing combines task-specific
advantage functions during training \cite{fu2022deep}, cooperative policy
learning coordinates separately parameterized locomotion and manipulation
policies \cite{pan2025roboduet}, and multi-critic learning separates reward
signals to reduce interference between objectives \cite{vijayan2025multi}.
Despite these advances, consistently reconciling precise end-effector
tracking with robust locomotion across diverse task conditions remains an
open problem.

Reusing the locomotion policy through an explicit dynamics model offers a
promising way to retain its mobility capability while handling manipulation
objectives.
Offline approaches identify low-dimensional linear models of RL-controlled
robots \cite{li2022bridging} or learn motion models that map high-level
commands to the response of a black-box locomotion controller
\cite{taouil2023blackbox}.
Online dynamics learning further adapts predictions during deployment.
For example, representative methods use random Fourier features (RFFs) for recursive
identification \cite{zhou2025simultaneous} or dual Gaussian processes to
combine rapid adaptation with long-term memory \cite{liu2025learning}.
However, these methods assume homogeneous system dynamics, where each
observed state corresponds to a single response.
They therefore pool observations from hidden operating conditions without
distinguishing their conditional responses.
This is restrictive when the same observed state and command correspond to
different response distributions.
Changes in the proportions of these conditions shift the pooled mean, while
differences among their conditional means introduce additional
between-condition covariance.
An offline model will consequently lose accuracy when the deployment
conditions differ from those represented in its identification data.
Online updates will follow transient mean shifts and alter the reference
response needed for subsequent compensation.
With limited online data, prediction errors also reflect epistemic
uncertainty in the learned mean, complicating the distinction between an
informative sample and a deviation associated with another operating
condition \cite{scampicchio2025gaussian}.
These effects reduce the reliability of learned predictions for precise
whole-body control.

To address this problem, an online learning method is proposed to account
for heterogeneous locomotion dynamics under a frozen reinforcement learning
policy and enable precise loco-manipulation.
The learner targets a reference residual mean associated with routine task
dynamics, including whole-body coupling and sustained task contact.
A contaminated Gaussian model anchors both components to the reference mean
and approximates additional deviations associated with other latent
conditions through covariance inflation.
A Gaussian process constructed with RFFs estimates the reference mean and
quantifies epistemic uncertainty.
Posterior probabilities from the contaminated Gaussian model account for
the epistemic uncertainty when weighting online updates, limiting the
influence of heterogeneous deviations while supporting learning in poorly
observed regions.
The learned residual model is then integrated into whole-body model
predictive control (MPC), where epistemic uncertainty regulates feedforward
compensation along the prediction horizon, limiting reliance on unreliable
predictions in regions insufficiently covered by online data.
As shown in Fig.~\ref{fig:hardware_show},
the resulting method accounts for heterogeneous responses, enabling precise loco-manipulation without
retraining the locomotion policy.

The main contributions of this work are summarized as follows:
\begin{enumerate}
    \item The statistical heterogeneity of locomotion dynamics under a frozen
    RL policy is characterized through latent-conditioned response
    distributions. The effects of pooled mean shifts and between-condition
    covariance are analyzed, motivating a contaminated Gaussian approximation
    centered on a task-relevant reference mean.

    \item A posterior-weighted RFF online learner is proposed to account
    jointly for heterogeneous response deviations and epistemic uncertainty
    when learning locomotion dynamics under a frozen policy. A finite-sample
    error bound is established to ensure uniform convergence in online
    learning.

    \item A residual-augmented MPC controller is developed to
    regulate learned feedforward compensation using epistemic uncertainty
    along the prediction horizon. Simulations across manipulator
    configurations and hardware experiments involving trajectory tracking,
    object grasping, door opening, disturbance recovery, and manipulation on
    the move demonstrate precise loco-manipulation without retraining the
    locomotion policy.
\end{enumerate}

\section{Related Work}\label{sec:related_work}
\subsection{Whole-Body Control for Loco-Manipulation}
Model-based control coordinates floating-base, manipulator, and contact motions through
online optimization of task objectives and physical constraints, allowing
task priorities to change without policy retraining.
Unified formulations combine centroidal and object dynamics for end-effector
tracking and door opening under prescribed contact schedules
\cite{sleiman2021unified}, while robust trajectory optimization incorporates
disturbance rejection \cite{ferrolho2023roloma}.
However, errors in robot dynamics, object properties, and contact models can
weaken prediction accuracy and execution robustness.

Learning-based control acquires coordinated responses from sampled
interactions without analytically describing every mechanism.
Examples include hierarchical visual policies for locomotion and grasping
\cite{liu2025visual} and whole-body controllers tracking task-frame
trajectories from demonstration-trained policies \cite{ha2025umionlegs}.
Training across varied conditions supports robust behaviors, but precise
end-effector regulation with stable locomotion requires sufficient sampling
near desired behaviors and balanced task objectives.
Additional task requirements or environmental conditions will demand retraining.

Hybrid approaches combine learned robustness with online optimization by
conditioning locomotion policies on MPC-predicted interaction wrenches
\cite{ma2022combining} or augmenting model-based feedforward control with RL
feedback
\cite{cheng2025rambo}.
Their integration requires a closed-loop model of the policy, its
observations, and environment interaction, which is difficult to
derive directly from rigid-body equations.
Accurately predicting this response for online task optimization across
changing conditions remains challenging for precise loco-manipulation.

\subsection{Heterogeneous Dynamics and Dynamics Learning}
In statistical modeling, heterogeneity describes systematic differences
among populations or data-generating conditions.
Latent-variable models represent such differences when the underlying
conditions are unobserved \cite{verbeke2017latent}.
The corresponding response distributions can differ in mean, covariance,
or both, even for the same observed inputs.
For robot dynamics, these conditions can include unobserved contact states
or changes in the environment.
Related dynamics models account for these condition-dependent differences
explicitly.
Hidden-parameter Markov decision processes use latent factors to describe
families of related systems \cite{doshivelez2016hidden}.
Gaussian-process models of switching port-Hamiltonian systems jointly learn
dynamics and switching conditions \cite{beckers2023switching}.
These approaches motivate modeling the effects of hidden conditions when
learning policy-controlled dynamics.

For dynamics learning, offline models predict controller responses from
identification data \cite{li2022bridging,taouil2023blackbox} and support
safety filtering \cite{yang2025shield}.
Online approaches use nonparametric adaptation \cite{boffi2022nonparametric},
adaptive force compensation \cite{sombolestan2024adaptive}, recursive
identification \cite{zhou2025simultaneous}, or neural dynamics updates
\cite{jiahao2023online}.
Gaussian-process methods combine rapid adaptation with long-term memory
\cite{liu2025learning}, while Bayesian multi-task models share information
across related manipulation tasks \cite{arcari2023bayesian}.
Hidden-condition response differences can still affect these estimates.

Dynamics learning must also assess epistemic uncertainty, which reflects
limited data and incomplete knowledge of the model
\cite{depeweg2018decomposition}.
Bayesian models quantify this uncertainty, and GP-based MPC propagates it
to assess prediction reliability.
During MPC rollouts, predictions can enter regions with limited data,
reducing model reliability over the remaining prediction horizon
\cite{hewing2020cautious}.
Large prediction errors can arise from either incomplete learning or
hidden-condition changes.
Treating both sources alike can reject informative observations or distort
model updates.
Distinguishing epistemic uncertainty from response heterogeneity while
assessing prediction reliability therefore remains difficult.

\section{Problem Statement}\label{sec:problem_statement}
Consider a quadruped manipulator with a frozen pretrained locomotion policy
$\bm{a}_{\ell,k}=\bm{\pi}_{\mathrm{loco}}(\bm{o}_k,\bm{c}_{\ell,k})$,
where $\bm{o}_k\in\mathbb{R}^{n_o}$ is the policy observation and
$\bm{a}_{\ell,k}$ denotes the leg-joint action. The controller selects
$\bm{u}_k=[\bm{c}_{\ell,k}^{\top}\ \bm{u}_{a,k}^{\top}]^{\top}$, where
$\bm{c}_{\ell,k}=[v_{x,k}\ v_{y,k}\ \omega_{z,k}\ \theta_k^{\mathrm{d}}]^{\top}
\in\mathbb{R}^{n_c}$, $n_c=4$, specifies body-frame planar velocities, yaw rate, and
base pitch, while $\bm{u}_{a,k}=\dot{\bm{q}}_{a,k}$ commands the manipulator
joint velocities. Thus, floating-base motion is regulated through the policy
command space, while the manipulator is commanded directly.

\begin{figure}[!t]
\centering
\includegraphics[width=0.85\columnwidth]{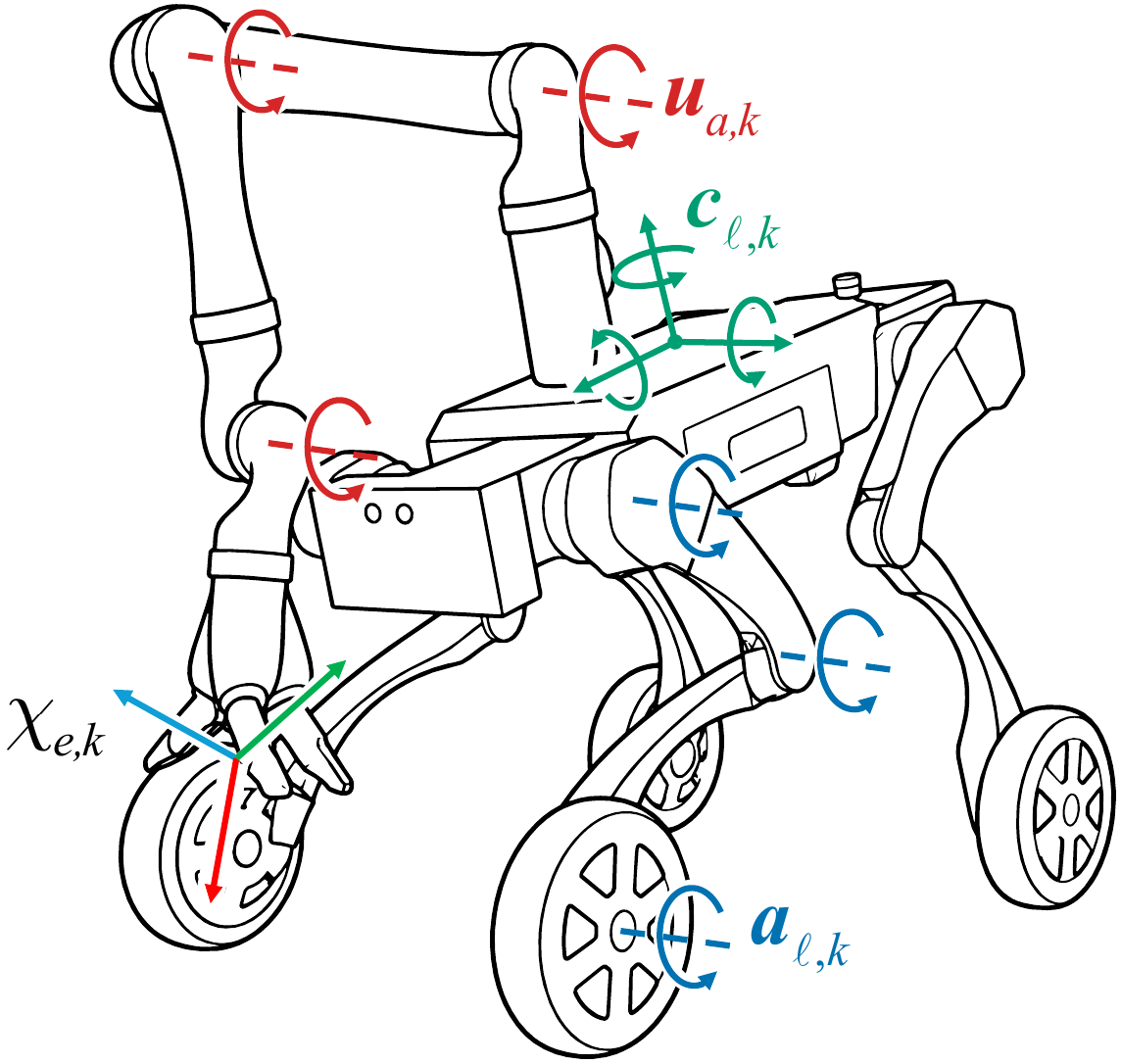}
\caption{
Control structure of the proposed method.
The control inputs are the locomotion command
$\bm{c}_{\ell,k}$ (green) 
and the manipulator joint-velocity command $\bm{u}_{a,k}$ (red).
Locomotion commands are specified within the frozen policy's
command space and mapped to leg-joint actions $\bm{a}_{\ell,k}$
(blue), while manipulator commands are applied directly.
The coordinated base and manipulator motion regulates the
end-effector pose $\bm{\chi}_{e,k}$.
}
\label{fig:control_architecture}
\end{figure}

Omitting leg-joint states, the reduced-order state is
$\bm{x}_k=[\bm{x}_{b,k}^{\top}\ \bm{q}_{a,k}^{\top}]^{\top}
\in\mathbb{R}^{6+n_a}$, with
$\bm{x}_{b,k}=[\bm{p}_{b,k}^{\top}\ \bm{\Phi}_{b,k}^{\top}]^{\top}$, where
$\bm{p}_{b,k}=[p_{x,k}\ p_{y,k}\ z_k]^{\top}$,
$\bm{\Phi}_{b,k}=[\psi_k\ \theta_k\ \phi_k]^{\top}$ are the base position
and yaw--pitch--roll angles, and $\bm{q}_{a,k}\in\mathbb{R}^{n_a}$ contains
the manipulator joint angles. The position--quaternion end-effector pose
and its error relative to the desired pose $\bm{\chi}_{e,k}^{\mathrm{ref}}$ are
\begin{align}
\bm{\chi}_{e,k}&=\bm{h}_e(\bm{x}_k)\in\mathbb{R}^{3}\times\mathbb{S}^{3},
\label{eq:end_effector_forward_kinematics}\\
\bm{e}_{e,k}&=\bm{\chi}_{e,k}\boxminus\bm{\chi}_{e,k}^{\mathrm{ref}}
\in\mathbb{R}^{6},
\label{eq:end_effector_tracking_error}
\end{align}
where $\bm{h}_e$ is the forward-kinematics map~\cite{mishra2023reduced},
and $\boxminus$ combines position difference with minimal quaternion
orientation error~\cite{bloesch2016primer}.
As shown in Fig.~\ref{fig:control_architecture}, 
the robot state $\bm{x}_k$ is regulated through the input $\bm{u}_k$.
Specifically, the frozen policy maps $\bm{o}_k$ and $\bm{c}_{\ell,k}$ to the
leg action $\bm{a}_{\ell,k}$ that governs the floating-base motion, while
$\bm{u}_{a,k}$ commands the manipulator motion. The objective is to select
$\bm{u}_k$ such that $\bm{e}_{e,k}$ is minimized while keeping
$\bm{\pi}_{\mathrm{loco}}$ frozen. To this end, a dynamics model is required
to capture the robot-state evolution from $\bm{x}_k$ to $\bm{x}_{k+1}$ under
the command $\bm{u}_k$.

\begin{figure*}[!t]
\centering
{\footnotesize
\begin{tabular}{@{}*{5}{l@{\hspace{1em}}}l@{}}
\residuallegendmark{residualZero}{}{circle}
\enspace $F_y=0\,\mathrm{N}$ &
\residuallegendmark{residualThirty}
  {dash pattern=on 4pt off 2.5pt}{rectangle}
\enspace $F_y=30\,\mathrm{N}$ &
\residuallegendmark{residualSixty}
  {dash pattern=on 0.9pt off 1.8pt}
  {regular polygon,regular polygon sides=3,minimum size=0.85em}
\enspace $F_y=60\,\mathrm{N}$ &
\residuallegendmark{residualPooled}{draw=none}
  {star,star points=5,star point ratio=2.3,minimum size=1em}
\enspace Pooled mean  &
\residualcovlegend{residualSigmaZero}{}
\enspace Reference covariance ${\Sigma}_0$ &
\residualcovlegend{residualSigmaInflated}
  {dash pattern=on 4pt off 2.5pt}
\enspace Inflated covariance $\kappa{\Sigma}_0$
\end{tabular}\par
}
\vspace{0.45em}
\subfloat[Homogeneous residual distribution.\label{fig:heteroscedastic_dynamics_pooled}]{%
\includegraphics[width=0.32\textwidth]{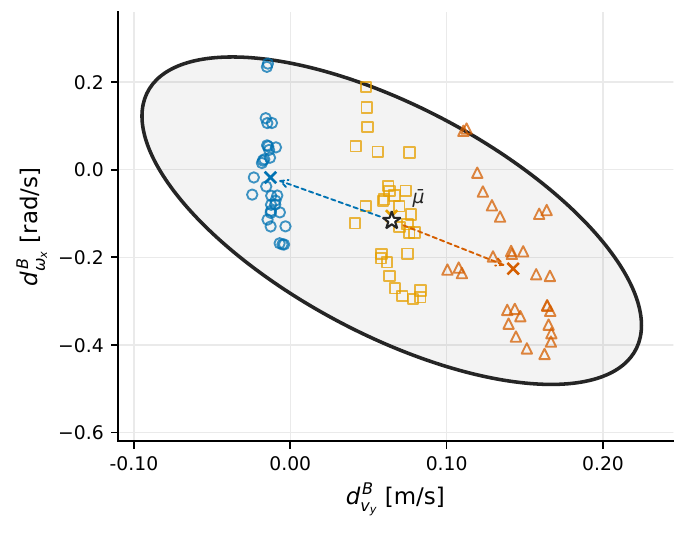}}
\hfill
\subfloat[Heterogeneous residual distributions.\label{fig:heteroscedastic_dynamics_by_force}]{%
\includegraphics[width=0.32\textwidth]{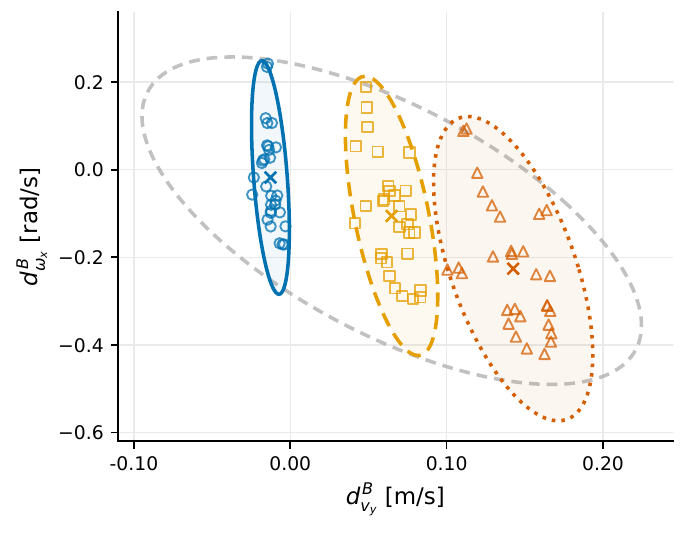}}
\hfill
\subfloat[Contaminated Gaussian model.\label{fig:heteroscedastic_dynamics_centered}]{%
\includegraphics[width=0.32\textwidth]{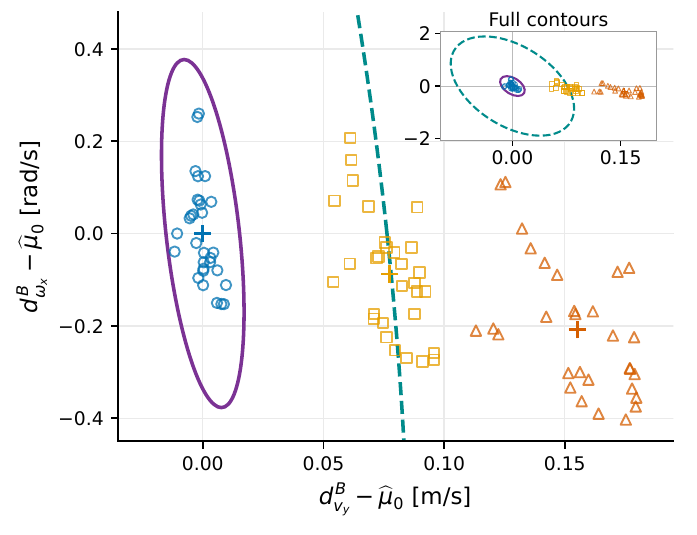}}
\caption{Heterogeneous locomotion dynamics illustrated by residuals of the
base lateral velocity $v_y$ and roll rate $\omega_x$ under lateral
forces $F_y=0$, $30$, and $60$~N.
Blue circles, orange squares, and vermilion triangles denote samples
under these force levels, respectively. Matching cross markers
indicate the corresponding group means.
(a) A homogeneous fit to pooled residuals places the mean between
latent-conditioned responses and combines within-condition variability
with between-condition differences.
The black star and shaded ellipse denote the pooled mean and covariance
region, respectively, while colored dashed arrows connect the pooled
mean to the group means.
(b) Heterogeneous residual distributions exhibit distinct means
and within-condition covariances under different latent conditions.
The solid blue, dashed orange, and dotted vermilion ellipses and their
shaded interiors represent the corresponding within-condition covariance
regions, while the gray dashed ellipse shows the pooled covariance
for comparison.
(c) The proposed contaminated Gaussian approximation uses two
components sharing the reference mean of the $F_y=0$~N condition,
with reference covariance $\Sigma_0$ and inflated covariance
$\kappa\Sigma_0$ to account for additional dispersion.
The solid purple and dashed teal ellipses denote $\Sigma_0$ and
$\kappa\Sigma_0$, respectively. 
}
\label{fig:heteroscedastic_dynamics}
\end{figure*}

The robot dynamics are modeled as
\begin{equation}
\bm{x}_{k+1}=\bm{f}_{\mathrm{nom}}(\bm{x}_k,\bm{u}_k)+\bm{B}_d\bm{d}_k,
\label{eq:nominal_dynamics}
\end{equation}
where $\bm{d}_k\in\mathbb{R}^{6}$ captures the discrepancy in floating-base
motion under the frozen policy, and
$\bm{B}_d=[\bm{I}_6\ \bm{0}_{6\times n_a}]^{\top}$ restricts residual
correction to the base. With sampling period $\Delta t$, the nominal model is
\begin{equation}
\bm{f}_{\mathrm{nom}}(\bm{x}_k,\bm{u}_k)
=
\begin{bmatrix}
p_{x,k}
+
\Delta t
(\cos\psi_k v_{x,k}
-\sin\psi_k v_{y,k})
\\[1mm]
p_{y,k}
+
\Delta t
(\sin\psi_k v_{x,k}
+\cos\psi_k v_{y,k})
\\[1mm]
z_k
+
\Delta t
\dfrac{z^{d}-z_k}{\tau_z}
\\[1mm]
\psi_k
+
\Delta t\omega_{z,k}
\\[1mm]
\theta_k
+
\Delta t
\dfrac{\theta_k^{\mathrm{d}}-\theta_k}{\tau_\theta}
\\[1mm]
\phi_k
-
\Delta t\dfrac{\phi_k}{\tau_\phi}
\\[1mm]
\bm{q}_{a,k}
+
\Delta t\dot{\bm{q}}_{a,k}
\end{bmatrix},
\label{eq:nominal_wholebody}
\end{equation}
Here, $z^d$ denotes the desired base height, and $\tau_z$, $\tau_\theta$, 
and $\tau_\phi$ are the response time constants for nominal 
height, pitch, and roll regulation, respectively.

The residual $\bm{d}_k$ inherits the partial observability of the POMDP
underlying the locomotion policy $\bm{\pi}_{\mathrm{loco}}$. Let
$\bm{\xi}_k\in\mathcal{Z}$ denote the unobserved latent state associated
with, for example, contact or environmental conditions. At transition $k$,
the robot state $\bm{x}_k$, policy observation $\bm{o}_k$, and applied
command $\bm{u}_k$ are available. The information is summarized as
$\bm{s}_k=[\bm{x}_k^{\top}\ \bm{o}_k^{\top}\ \bm{u}_k^{\top}]^{\top}
\in\mathbb{R}^{n_s}$, where $n_s=6+2n_a+n_o+n_c$. Because $\bm{s}_k$
provides only a partial observation of the underlying system, it does not
uniquely determine $\bm{\xi}_k$. The remaining uncertainty is described by
the conditional distribution $p(\bm{\xi}_k\mid\bm{s}_k)$ over latent states
compatible with $\bm{s}_k$. Suppressing the transition index, the
observable residual distribution is therefore obtained by marginalizing
the latent-conditioned responses with respect to $p(\bm{\xi}\mid\bm{s})$.
\begin{equation}
p(\bm{d}\mid\bm{s})=\int_{\mathcal{Z}}
p(\bm{d}\mid\bm{\xi},\bm{s})p(\bm{\xi}\mid\bm{s})\,d\bm{\xi}.
\label{eq:latent_marginalized_residual}
\end{equation}
In~\eqref{eq:latent_marginalized_residual}, a given $\bm{s}$ can correspond
to different latent states whose conditional residual distributions
$p(\bm{d}\mid\bm{\xi},\bm{s})$ differ in mean, covariance, or both,
giving rise to heterogeneous residual responses. These heterogeneous
responses are mixed according to $p(\bm{\xi}\mid\bm{s})$ under the same
observable information. The laws of total expectation and total covariance
give the conditional moments of $p(\bm{d}\mid\bm{s})$ as
\begin{subequations}
\label{eq:marginal_residual_moments}
\begin{align}
\mathbb{E}[\bm{d}\mid\bm{s}]
&=
\int_{\mathcal{Z}}
\mathbb{E}[\bm{d}\mid\bm{\xi},\bm{s}]
p(\bm{\xi}\mid\bm{s})
\,d\bm{\xi},
\label{eq:marginal_residual_mean}\\
\operatorname{Cov}(\bm{d}\mid\bm{s})
&=
\mathbb{E}_{\bm{\xi}\mid\bm{s}}
\bigl[\operatorname{Cov}(\bm{d}\mid\bm{\xi},\bm{s})\bigr]
+
\operatorname{Cov}_{\bm{\xi}\mid\bm{s}}
\bigl[\mathbb{E}[\bm{d}\mid\bm{\xi},\bm{s}]\bigr].
\label{eq:marginal_residual_covariance}
\end{align}
\end{subequations}

The marginal mean in~\eqref{eq:marginal_residual_mean} averages the
conditional means according to their occurrence probabilities. Separated
responses can produce a multimodal distribution whose mean lies between
modes, away from typical responses under individual conditions. A pooled
mean can therefore misrepresent the response needed for compensation
under a particular condition.
Fig.~\ref{fig:heteroscedastic_dynamics_pooled} illustrates this effect using
base lateral-velocity and roll-rate residuals under lateral forces
$F_y=0$, $30$, and $60$~N. The black star lies between the separated
force-group means, and its location depends on their sample proportions.
Changes in those proportions can shift the estimated mean even when the
conditional response of each group remains unchanged.

Equation~\eqref{eq:marginal_residual_covariance} separates the pooled
covariance into two contributions. The first averages the variability
within each latent condition. The second describes differences among
conditional means and adds a positive semidefinite between-condition
covariance. The overall dispersion therefore increases relative to the
average within-condition covariance.
In Fig.~\ref{fig:heteroscedastic_dynamics_by_force}, the colored ellipses
describe within-group variability, while the separation among group means
indicates between-condition differences. The black ellipse in
Fig.~\ref{fig:heteroscedastic_dynamics_pooled} and the gray dashed ellipse in
Fig.~\ref{fig:heteroscedastic_dynamics_by_force} combine both contributions.
The pooled covariance consequently need not represent the variability
under an individual condition.

With limited online data and no latent-condition labels, fluctuations in
sampled condition proportions perturb mean estimates, while the
additional dispersion increases their sampling variability. Samples with
distinct conditional means can pull successive updates toward different
responses, affecting prediction and compensation. These
prediction errors propagate through $\bm{h}_e$ to end-effector pose errors.
Online learning should therefore account for heterogeneous responses to
support precise loco-manipulation under the frozen policy.

\begin{problem}[\mdseries Online Learning of Heterogeneous Dynamics for Precise Loco-Manipulation]
Given the frozen locomotion policy $\bm{\pi}_{\mathrm{loco}}$, suppose that the current robot state $\bm{x}_k$, previous policy observation
$\bm{o}_{k-1}$, and the previous state--input pair
$(\bm{x}_{k-1},\bm{u}_{k-1})$ are available.
At each transition $k=1,\ldots,T$, learn the unknown heterogeneous residual $\bm{d}_k$ in \eqref{eq:nominal_dynamics}
online and determine the optimal command $\bm{u}_k^{\star}$ subject to the
forward-kinematics model in
\eqref{eq:end_effector_forward_kinematics} and the dynamics model in
\eqref{eq:nominal_dynamics}, such that the end-effector tracking error
$\bm{e}_{e,k}$ in \eqref{eq:end_effector_tracking_error} is minimized.
\end{problem}

\section{Method}\label{sec:method}
\begin{figure*}[!t]
\centering
\includegraphics[width=\textwidth]{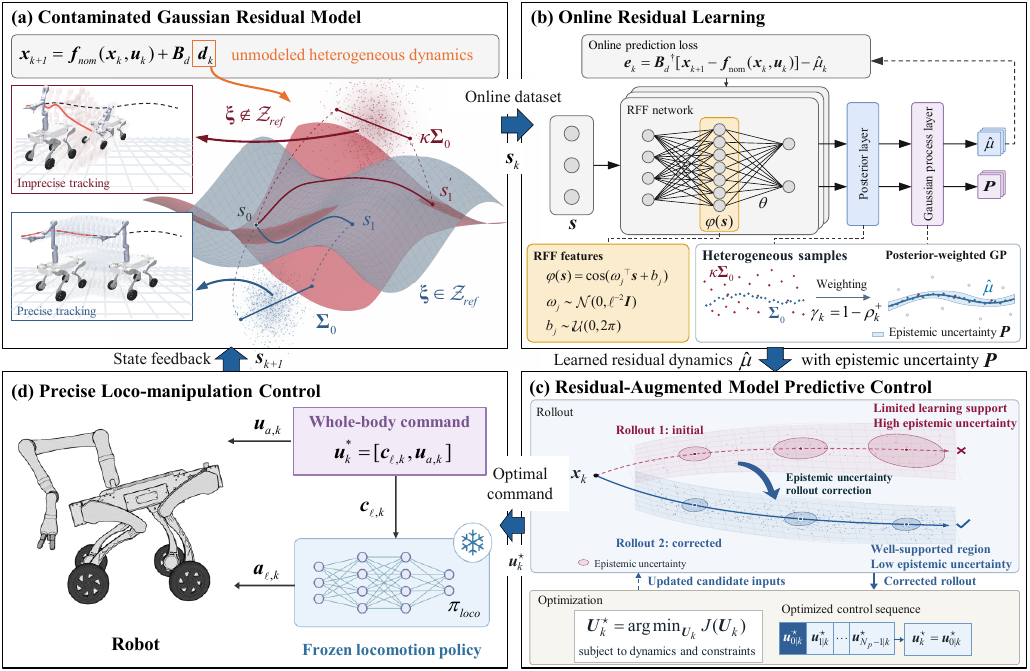}
\caption{Overall architecture of the proposed method.
(a) The contaminated Gaussian residual model approximates heterogeneous
responses using a shared reference mean and two covariance scales,
$\Sigma_0$ and $\kappa\Sigma_0$.
The blue and red distributions represent reference and nonreference
responses, respectively.
(b) The online learner uses an RFF-based Gaussian process to estimate
the reference residual mean and quantify epistemic uncertainty.
Posterior learnability probabilities account for this uncertainty
when weighting model updates, limiting the influence of nonreference
responses.
The blue curve and shaded band illustrate the learned mean
and epistemic uncertainty, respectively.
(c) Residual-augmented MPC rolls out the learned residual mean,
epistemic covariance, and regime probability over the prediction
horizon. The predicted epistemic uncertainty corrects the rollout by attenuating
residual compensation, limiting MPC's reliance on unreliable residual
predictions in insufficiently learned regions.
The red and blue trajectories illustrate the initial and corrected
rollouts, respectively, with ellipses indicating epistemic uncertainty.
(d) The optimized whole-body command combines locomotion commands
mapped to leg-joint actions by the frozen policy and manipulator
joint-velocity commands applied directly to the robot.
}
\label{fig:method_framework}
\end{figure*}

\subsection{Contaminated Gaussian Residual Model}\label{sec:probabilistic_residual}
The online learner targets a repeatable residual response for precise
state prediction and compensation. However, the marginalization in
\eqref{eq:latent_marginalized_residual} mixes heterogeneous responses
associated with unobserved latent conditions, potentially shifting the
marginal mean away from the task-relevant response while adding
between-condition dispersion, as described by
\eqref{eq:marginal_residual_moments}. To address these effects without
resolving the continuum of latent states, the latent-state integral in \eqref{eq:latent_marginalized_residual} is
approximated by a mixture of two latent components distinguished relative
to the following reference operating conditions.

\begin{definition}[\mdseries Reference Operating Conditions]
\label{def:reference_operating_conditions}
The reference operating conditions are defined as a set
$\mathcal{Z}_{\mathrm{ref}}\subseteq\mathcal{Z}$ of latent states
associated with routine task dynamics, including whole-body coupling
and the effects of sustained contact forces required by the task.
For observable information $\bm{s}$ satisfying
$\Pr(\bm{\xi}\in\mathcal{Z}_{\mathrm{ref}}\mid\bm{s})>0$,
the corresponding conditional residual mean is defined as
\begin{equation}
\bm{\mu}(\bm{s})
:=
\mathbb{E}\!\left[
\bm{d}\mid\bm{s},\bm{\xi}\in\mathcal{Z}_{\mathrm{ref}}
\right],
\qquad
\bm{\mu}_k:=\bm{\mu}(\bm{s}_k).
\label{eq:reference_residual_mean}
\end{equation}
\end{definition}

Equation~\eqref{eq:marginal_residual_covariance} shows that differences
among latent-conditioned means contribute to the overall residual
covariance. When measured relative to the mean $\bm{\mu}(\bm{s})$ of
the reference operating conditions
$\bm{\xi}\in\mathcal{Z}_{\mathrm{ref}}$, residual responses under
$\bm{\xi}\notin\mathcal{Z}_{\mathrm{ref}}$ can exhibit additional
deviations due to their conditional mean shifts. A covariance-scale
distinction between these latent regimes is therefore adopted in the
following definition.

\begin{definition}[\mdseries Residual Learnability]
\label{def:residual_learnability}
A binary latent learnability variable $\zeta_k\in\{0,1\}$ is defined
according to whether the latent state $\bm{\xi}_k$ belongs to the reference operating conditions
$\mathcal{Z}_{\mathrm{ref}}$. The corresponding Gaussian covariance
parameters are assigned as
\begin{subequations}
\label{eq:learnability_regime_covariances}
\begin{align}
\bm{\xi}_k\in\mathcal{Z}_{\mathrm{ref}}:\quad
&\zeta_k=0,\quad
\bm{\Sigma}_{d,k}^{(0)}=\bm{\Sigma}_0,
\label{eq:baseline_latent_mode}\\
\bm{\xi}_k\notin\mathcal{Z}_{\mathrm{ref}}:\quad
&\zeta_k=1,\quad
\bm{\Sigma}_{d,k}^{(1)}=\kappa\bm{\Sigma}_0.
\label{eq:inflated_latent_mode}
\end{align}
\end{subequations}
The regimes $\zeta_k=0$ and $\zeta_k=1$ are termed \emph{learnability}
and \emph{unlearnability}, respectively. 
\end{definition}

In Definition~\ref{def:residual_learnability},
$\bm{\Sigma}_{d,k}^{(\zeta_k)}$ denotes the covariance assigned to regime
$\zeta_k$, with the given matrix $\bm{\Sigma}_0\succ\bm{0}$ approximating
$\operatorname{Cov}(\bm{d}_k\mid\bm{s}_k,
\bm{\xi}_k\in\mathcal{Z}_{\mathrm{ref}})$, while $\kappa>1$ provides an
inflated scale for departures from the reference response.
To address the mixture-induced mean shift in
\eqref{eq:marginal_residual_mean}, both Gaussian components are centered
on the reference mean $\bm{\mu}(\bm{s})$ in
Definition~\ref{def:reference_operating_conditions}, keeping the
predictive mean of the approximation independent of the component
proportions. The additional dispersion induced by the between-condition
covariance in \eqref{eq:marginal_residual_covariance} is approximated
using the inflated covariance rather than absorbed into the reference
covariance. 
Under Definitions~\ref{def:reference_operating_conditions}
and~\ref{def:residual_learnability}, the resulting regime-conditioned
Gaussian approximation is expressed as
\begin{equation}
q(\bm{d}_k\mid\bm{s}_k,\zeta_k)
=
\mathcal{N}\!\left(
\bm{d}_k,
\bm{\mu}_k,
\kappa^{\zeta_k}\bm{\Sigma}_0
\right).
\label{eq:conditional_residual_distribution}
\end{equation}
Marginalizing the binary latent variable $\zeta_k$ requires a mixing probability
that specifies the relative prevalence of the two regimes. To approximate
the occurrence probability of nonreference conditions without estimating
an additional state-dependent function from limited online data,
a given prior probability of the unlearnability regime is prescribed as
\begin{equation}
p_{\zeta_1}
:=
q(\zeta_k=1),
\qquad
0<p_{\zeta_1}<1.
\label{eq:unlearnability_prior}
\end{equation}
Marginalizing the binary latent variable
replaces the integral over $\mathcal{Z}$ in
\eqref{eq:latent_marginalized_residual} by the finite sum
\begin{gather}
q(\bm{d}_k\mid\bm{s}_k)
:=\sum_{j=0}^{1}q(\zeta_k=j)\,q(\bm{d}_k\mid\bm{s}_k,\zeta_k=j)
\nonumber\\
=(1-p_{\zeta_1})\mathcal{N}\!\left(\bm{d}_k,\bm{\mu}_k,\bm{\Sigma}_0\right)+p_{\zeta_1}\mathcal{N}\!\left(\bm{d}_k,\bm{\mu}_k,\kappa\bm{\Sigma}_0\right).
\label{eq:contaminated_gaussian_residual}
\end{gather}
Equation~\eqref{eq:contaminated_gaussian_residual} gives the resulting
contaminated Gaussian model for approximating the heterogeneous residual
distribution in \eqref{eq:latent_marginalized_residual}. 
Its first two conditional moments are
\begin{subequations}
\label{eq:contaminated_gaussian_moments}
\begin{align}
\mathbb{E}_q\!\left[
\bm{d}_k\mid\bm{s}_k
\right]
&=
\bm{\mu}_k,
\label{eq:contaminated_gaussian_mean}
\\
\operatorname{Cov}_q\!\left[
\bm{d}_k\mid\bm{s}_k
\right]
&=
\left[
1+(\kappa-1)p_{\zeta_1}
\right]\bm{\Sigma}_0.
\label{eq:contaminated_gaussian_covariance}
\end{align}
\end{subequations}
Equation~\eqref{eq:contaminated_gaussian_moments} gives a predictive mean
$\bm{\mu}_k$ independent of the mixture proportions and a covariance
$[1+(\kappa-1)p_{\zeta_1}]\bm{\Sigma}_0$.
Fig.~\ref{fig:heteroscedastic_dynamics_centered} illustrates the component
covariances $\bm{\Sigma}_0$ and $\kappa\bm{\Sigma}_0$ as concentric ellipses
centered on the $F_y=0$~N sample mean.

\subsection{Online Residual Learning}\label{subsec:rff_rls}
Following the analysis in Section~\ref{sec:probabilistic_residual},
the online learner models the residual response conditioned on the
current observable information $\bm{s}_k$ using
\eqref{eq:contaminated_gaussian_residual}.
Given the parameters $\bm{\Sigma}_0$, $\kappa$, and
$p_{\zeta_1}$, the objective is to learn the residual mean under reference operating conditions
$\bm{\mu}(\bm{s})$ in \eqref{eq:reference_residual_mean} online while probabilistically inferring whether
each incoming residual observation $\bm{d}_k$ originates from reference
or nonreference operating conditions.
The central challenge is that deviations from the predicted mean
reflect both residual variation and uncertainty in
the online learning process.
Condition inference should therefore account for the learner's epistemic
uncertainty rather than attributing large prediction errors solely
to nonreference responses.

\begin{assumption}[\mdseries RKHS Regularity of the Conditional Residual Mean]
\label{ass:residual_mean_rkhs}
Each component of the residual mean under reference operating conditions
$\bm{\mu}:\mathbb{R}^{n_s}\rightarrow\mathbb{R}^{n_d}$ in
\eqref{eq:contaminated_gaussian_residual} belongs to the RKHS
$\mathcal{H}_{\mathcal{K}}$ associated with a known kernel
$\mathcal{K}$, i.e.,
\begin{equation}
\mu_j(\cdot)\in\mathcal{H}_{\mathcal{K}},
\qquad
\|\mu_j\|_{\mathcal{H}_{\mathcal{K}}}
\le B_{\mu},
\qquad
j=1,\ldots,n_d,
\label{eq:mean_rkhs_assumption}
\end{equation}
where $B_{\mu}>0$ is finite and $n_d=6$.
\end{assumption}

Under Assumption~\ref{ass:residual_mean_rkhs}, the RFF feature vector is defined as
\begin{equation}
\bm{\phi}(\bm{s})
=
\sqrt{\frac{2}{D}}
\begin{bmatrix}
\cos(\bm{\omega}_1^{\top}\bm{s}+b_1)\\
\vdots\\
\cos(\bm{\omega}_D^{\top}\bm{s}+b_D)
\end{bmatrix}.
\label{eq:rff_map}
\end{equation}
For the Gaussian kernel with length scale $\ell>0$, the random frequencies
and phases are sampled as
$\bm{\omega}_j\sim\mathcal{N}(\bm{0},\ell^{-2}\bm{I}_{n_s})$ and
$b_j\sim\mathcal{U}(0,2\pi)$, respectively, for $j=1,\ldots,D$.
Consequently, the conditional mean
$\bm{\mu}(\bm{s})$ is represented in finite-dimensional form as
\begin{equation}
\bm{\mu}(\bm{s})
=
\bm{W}\bm{\phi}(\bm{s}).
\label{eq:rff_residual_mean_model}
\end{equation}
The equality in \eqref{eq:rff_residual_mean_model} is adopted by
neglecting the finite-dimensional RFF representation error \cite{rahimi2008uniform}.
Here, $\bm{W}\in\mathbb{R}^{D\times n_d}$ denotes the unknown RFF
parameter matrix. Define the vectorized parameter and its design matrix as
\begin{subequations}
\label{eq:vectorized_rff_model}
\begin{align}
\bm{\theta}
&:=\operatorname{vec}(\bm{W})
\in\mathbb{R}^{Dn_d},
\label{eq:vectorized_rff_parameter}\\
\bm{H}(\bm{s})
&:=
\bm{I}_{n_d}\otimes\bm{\phi}(\bm{s})^\top
\in\mathbb{R}^{n_d\times Dn_d}.
\label{eq:rff_design_matrix}
\end{align}
\end{subequations}
At transition $k$, the online learning dataset is defined as
\begin{equation}
\mathcal{D}_k
:=
\{(\bm{s}_i,\bm{d}_i)\}_{i=0}^{k-1}.
\label{eq:bayesian_rff_dataset}
\end{equation}
It contains the observable information $\bm{s}_i$ paired with the
residual observation $\bm{d}_i$ from each completed transition before
the current residual $\bm{d}_k$ is observed.

To quantify epistemic uncertainty in the learned results, the RFF weight vector
$\bm{\theta}$ is represented by a Gaussian posterior approximation
rather than a deterministic point estimate as
\begin{subequations}
\label{eq:bayesian_rff_parameter_posterior}
\begin{align}
p(\bm{\theta}\mid\mathcal{D}_k)
&=
\mathcal{N}(\bm{\theta},\bm{\theta}_k,\bm{P}_k),
\label{eq:bayesian_rff_posterior_distribution}\\
\bm{\theta}_k
&:=
\mathbb{E}[\bm{\theta}\mid\mathcal{D}_k]
\in\mathbb{R}^{Dn_d},
\label{eq:bayesian_rff_parameter_mean}\\
\bm{P}_k
&:=
\operatorname{Cov}[\bm{\theta}\mid\mathcal{D}_k]
\in\mathbb{S}_{++}^{Dn_d}.
\label{eq:bayesian_rff_parameter_covariance}
\end{align}
\end{subequations}
Here, $\bm{\theta}_k$ provides the current parameter estimate, while
$\bm{P}_k$ quantifies its epistemic uncertainty conditioned on
$\mathcal{D}_k$. 
Consequently, the posterior distribution of the conditional mean is
\begin{equation}
\bm{\mu}(\bm{s})\mid\mathcal{D}_k
\sim
\mathcal{N}\!\left(
\bm{H}(\bm{s})\bm{\theta}_k,\,
\bm{H}(\bm{s})\bm{P}_k\bm{H}(\bm{s})^\top
\right).
\label{eq:bayesian_rff_mean_posterior}
\end{equation}


At transition $k$, let
$\bm{H}_k:=\bm{H}(\bm{s}_k)$ and define the conditional-mean prediction as
\begin{equation}
\hat{\bm{\mu}}_k
=
\bm{H}_k\bm{\theta}_k.
\label{eq:residual_mean_learning}
\end{equation}
Each measured transition yields the residual observation
\begin{equation}
\bm{d}_k
=
\bm{B}_d^{\dagger}
\left[
\bm{x}_{k+1}
-
\bm{f}_{\mathrm{nom}}(\bm{x}_{k},\bm{u}_{k})
\right],
\label{eq:measured_residual}
\end{equation}
where $\bm{B}_d^{\dagger}$ denotes the Moore--Penrose pseudoinverse of
$\bm{B}_d$. 
The corresponding innovation is
\begin{equation}
\bm{e}_k
=
\bm{d}_k-\hat{\bm{\mu}}_k.
\label{eq:residual_innovation}
\end{equation}
The innovation $\bm{e}_k$ contains both the irreducible residual variation and
the epistemic uncertainty arising from the online learning process.

Marginalizing the RFF parameters $\bm{\theta}^\star$ in
\eqref{eq:bayesian_rff_parameter_posterior} gives the epistemic covariance
\begin{equation}
\bm{A}_k
:=
\bm{H}_k\bm{P}_k\bm{H}_k^\top.
\label{eq:bayesian_rff_epistemic_covariance}
\end{equation}
Combining $\bm{A}_k$ with the residual variation $\kappa^{\zeta_k}\bm{\Sigma}_0$ in
\eqref{eq:contaminated_gaussian_residual} yields the predictive innovation
covariances
\begin{subequations}
\label{eq:predictive_regime_covariances}
\begin{align}
\bm{S}_{0,k}
&=
\bm{A}_k+\bm{\Sigma}_0,
\label{eq:learnability_predictive_covariance}\\
\bm{S}_{1,k}
&=
\bm{A}_k+\kappa\bm{\Sigma}_0.
\label{eq:unlearnability_predictive_covariance}
\end{align}
\end{subequations}
\begin{proposition}[\mdseries Posterior Unlearnability Probability]\label{prop:posterior_unlearnability}
Given the prior unlearnability probability $p_{\zeta_1}$,
the innovation $\bm{e}_k$ in \eqref{eq:residual_innovation}, and the
predictive covariance matrices $\bm{S}_{0,k},\bm{S}_{1,k}\succ\bm{0}$
in \eqref{eq:predictive_regime_covariances}, the posterior
unlearnability probability has the closed-form expression
\begin{equation}
\begin{aligned}
\rho_k^{+}
&:=
\Pr\!\left(\zeta_k=1\mid\bm{e}_k,\mathcal{D}_k\right)
\\
&=\sigma\!\left(
\begin{aligned}
&\log\frac{p_{\zeta_1}}{1-p_{\zeta_1}}
-\frac{1}{2}\log\frac{|\bm{S}_{1,k}|}{|\bm{S}_{0,k}|}\\
&+\frac{1}{2}\bm{e}_k^\top
\left(\bm{S}_{0,k}^{-1}-\bm{S}_{1,k}^{-1}\right)\bm{e}_k
\end{aligned}
\right).
\end{aligned}
\label{eq:posterior_unlearnability_probability}
\end{equation}
Here, $\sigma(z):=[1+\exp(-z)]^{-1}$ denotes the logistic sigmoid.
\end{proposition}

\begin{lemma}[\mdseries Sensitivity to Innovation and Epistemic Uncertainty]
\label{thm:residual_unlearnability_discriminability}
Given $\bm{\Sigma}_0\succ\bm{0}$, $\kappa>1$,
$0<p_{\zeta_1}<1$, and $\bm{A}_k\succeq\bm{0}$, the posterior
in \eqref{eq:posterior_unlearnability_probability} satisfies the
following properties.
\begin{subequations}
\label{eq:posterior_innovation_epistemic_limits}
(i) For uniformly bounded epistemic covariance $\bm{A}_k$,
the posterior approaches one as the normalized innovation
$\bm{\Sigma}_0^{-1/2}\bm{e}_k$ becomes unbounded:
\begin{equation}
\lim_{\|\bm{\Sigma}_0^{-1/2}\bm{e}_k\|_2\to\infty}
\rho_k^+=1.
\label{eq:posterior_innovation_ray_limit}
\end{equation}
(ii) For bounded normalized innovation squared 
$\bm{e}_k^\top\bm{S}_{0,k}^{-1}\bm{e}_k$ under the reference condition,
the posterior approaches the prior as epistemic uncertainty
dominates in every direction:
\begin{equation}
\lim_{\lambda_{\min}(
\bm{\Sigma}_0^{-1/2}\bm{A}_k\bm{\Sigma}_0^{-1/2}
)\to\infty}
\rho_k^+=p_{\zeta_1}.
\label{eq:posterior_epistemic_prior_limit}
\end{equation}
\end{subequations}
\end{lemma}
Lemma~\ref{thm:residual_unlearnability_discriminability} shows how
the prior contaminated Gaussian model in
\eqref{eq:contaminated_gaussian_residual} and the modeling of
epistemic covariance in \eqref{eq:bayesian_rff_mean_posterior}
help posterior condition inference distinguish evidence for
between-condition residual dispersion from uncertainty in the online
learning process. Innovations favor the nonreference
condition when epistemic covariance remains bounded, whereas dominant
epistemic uncertainty with bounded normalized innovation squared drives
the posterior toward the prior. 

With the posterior unlearnability probability, the posterior learnability 
probability is defined as
\begin{equation}
\gamma_k
:=
1-\rho_k^{+},
\label{eq:learnability_responsibility}
\end{equation}
which quantifies the posterior probability that the current residual belongs
to the reference operating conditions. Weighting the  Gaussian
log-likelihood by $\gamma_k$ yields the update gain for the
Gaussian parameter posterior in \eqref{eq:bayesian_rff_parameter_posterior},
\begin{equation}
\bm{K}_k
=
\gamma_k\bm{P}_k\bm{H}_k^\top
\left(
\bm{\Sigma}_0
+
\gamma_k\bm{H}_k\bm{P}_k\bm{H}_k^\top
\right)^{-1}.
\label{eq:posterior_weighted_rls_gain}
\end{equation}
The parameter mean and covariance are updated as
\begin{subequations}
\label{eq:posterior_weighted_rls_update}
\begin{align}
\bm{\theta}_{k+1}
&=
\bm{\theta}_k+\bm{K}_k\bm{e}_k,
\label{eq:posterior_weighted_weight_update}\\
\bm{P}_{k+1}
&=
\bm{P}_k-\bm{K}_k\bm{H}_k\bm{P}_k.
\label{eq:posterior_weighted_information_update}
\end{align}
\end{subequations}

\begin{lemma}[\mdseries Bounded and Redescending Behavior of Parameter Updates]
\label{lemma:bounded_redescending_influence}
Let
$\Delta\bm{\theta}_k:=\bm{\theta}_{k+1}-\bm{\theta}_k
=\bm{K}_k\bm{e}_k$.
For $\bm{P}_k\succ\bm{0}$, finite $\bm{H}_k$,
$\bm{\Sigma}_0\succ\bm{0}$, $\kappa>1$, and
$0<p_{\zeta_1}<1$, the parameter increment satisfies
\begin{subequations}
\label{eq:bounded_redescending_influence}
\begin{align}
\sup_{\bm{e}_k\in\mathbb{R}^{n_d}}
\|\Delta\bm{\theta}_k\|_2&<\infty,
\label{eq:bounded_parameter_increment}\\
\lim_{\|\bm{e}_k\|_2\rightarrow\infty}
\|\Delta\bm{\theta}_k\|_2&=0.
\label{eq:redescending_parameter_increment}
\end{align}
\end{subequations}
\end{lemma}
With Lemma~\ref{lemma:bounded_redescending_influence}, the
parameter update remains bounded and redescends to zero as
the innovation magnitude increases without bound. This property helps
protect the online learning process from the adverse effects of
between-condition covariance.

Algorithm~\ref{alg:online_residual_learning} summarizes the online
learning of the reference operating conditions using posterior learnability weights that
account for epistemic uncertainty.

\begin{theorem}[\mdseries Finite-Sample Online Learning Accuracy]
\label{thm:reference_mean_accuracy}
Assume that there is no RFF representation error in
\eqref{eq:rff_residual_mean_model} and that the  features
are persistently exciting.
Under the contaminated Gaussian model in
\eqref{eq:contaminated_gaussian_residual}, and the parameters updated
according to Algorithm~\ref{alg:online_residual_learning}, the online learning process satisfies
\begin{equation}
\sup_{\bm{s}\in \mathcal{D}}
\|\widehat{\bm{\mu}}_N(\bm{s})-\bm{\mu}(\bm{s})\|_2
\le\mathcal{O}\!\left(
\sqrt{\frac{\log(N+1)}{N+1}}
\right),
\label{eq:reference_mean_finite_sample_accuracy}
\end{equation}
where $N$ denotes the number of sampled transition pairs used for
online learning.
\end{theorem}

Theorem~\ref{thm:reference_mean_accuracy} gives a uniform error bound
for the online learning of the reference mean.
As the number of sampled transition pairs increases, the error bound approaches zero,
and the estimated mean $\widehat{\bm{\mu}}_N(\bm{s})$ converges to
the mean $\bm{\mu}(\bm{s})$.


\begin{algorithm}[t]
\caption{Online Residual Learning}\label{alg:online_residual_learning}
\begin{algorithmic}[1]
\REQUIRE Given $\mathcal{D}_k=\{(\bm{s}_i,\bm{d}_i)\}_{i=0}^{k-1}$,
$\bm{\Sigma}_0$, $\kappa$, and $p_{\zeta_1}$
\ENSURE $\hat{\bm{\mu}}_k$, $\rho_k^{+}$,
$\bm{\theta}_{k+1}$, and $\bm{P}_{k+1}$
\STATE Initialize
$\bm{\theta}_0=\bm{0}_{Dn_d}$ and
$\bm{P}_0=\bm{I}_{Dn_d}$
\FOR{$k=0,1,\ldots$}
\STATE Compute $\bm{\phi}(\bm{s}_k)$ and 
$\bm{H}_k$ using
\eqref{eq:rff_map} and \eqref{eq:rff_design_matrix}.
\STATE $\hat{\bm{\mu}}_k\gets\bm{H}_k\bm{\theta}_k$ \hfill
\eqref{eq:residual_mean_learning}
\STATE $\bm{A}_k\gets\bm{H}_k\bm{P}_k\bm{H}_k^\top$ \hfill
\eqref{eq:bayesian_rff_epistemic_covariance}
\STATE $\bm{S}_{0,k}\gets\bm{A}_k+\bm{\Sigma}_0$,
$\bm{S}_{1,k}\gets\bm{A}_k+\kappa\bm{\Sigma}_0$ \hfill
\eqref{eq:predictive_regime_covariances}
\STATE $\bm{e}_k\gets\bm{d}_k-\hat{\bm{\mu}}_k$ \hfill
\eqref{eq:residual_innovation}
\STATE Compute the posterior unlearnability probability
$\rho_k^{+}$ from $\bm{e}_k$, $\bm{S}_{0,k}$, and $\bm{S}_{1,k}$ using
\eqref{eq:posterior_unlearnability_probability}
\STATE $\gamma_k\gets1-\rho_k^{+}$ \hfill
\eqref{eq:learnability_responsibility}
\STATE $\bm{K}_k\gets
\gamma_k\bm{P}_k\bm{H}_k^\top
(\bm{\Sigma}_0+\gamma_k\bm{A}_k)^{-1}$
\hfill
\eqref{eq:posterior_weighted_rls_gain}
\STATE $\bm{\theta}_{k+1}\gets
\bm{\theta}_k+\bm{K}_k\bm{e}_k$ \hfill
\eqref{eq:posterior_weighted_weight_update}
\STATE $\bm{P}_{k+1}\gets
\bm{P}_k-\bm{K}_k\bm{H}_k\bm{P}_k$ \hfill
\eqref{eq:posterior_weighted_information_update}
\ENDFOR
\end{algorithmic}
\end{algorithm}

\subsection{Residual-Augmented Model Predictive Control}\label{subsec:residual_augmented_mpc}
At transition $k$, the learned residual model is characterized by the
RFF weight $\bm{\theta}_k$, the feature matrix $\bm{H}_k$,
the parameter posterior covariance $\bm{P}_k$, and the posterior
unlearnability probability $\rho_k^{+}$.
Achieving precise control using the learning results in
Section~\ref{subsec:rff_rls} involves rolling out the learned residual
model over the prediction horizon for feedforward compensation.
The effectiveness of this compensation is affected by
uncertainty introduced by online learning and model underfitting in
nonreference operating conditions.
Accounting for these issues helps prevent learning errors from amplifying
control errors and model underfitting from destabilizing the closed-loop system.

To this end, the residual mean, learning uncertainty, and
posterior unlearnability probability are rolled out over the
prediction horizon. At each prediction step $i$, these quantities are
evaluated as
\begin{subequations}
\label{eq:mpc_residual_rollout}
\begin{align}
\hat{\bm{\mu}}_{i\mid k}
&=
\bm{H}(\bm{s}_{i\mid k})\bm{\theta}_k,
\label{eq:mpc_residual_mean_rollout}\\
\bm{A}_{i\mid k}
&=
\bm{H}(\bm{s}_{i\mid k})\bm{P}_k
\bm{H}(\bm{s}_{i\mid k})^\top,
\label{eq:mpc_epistemic_covariance_rollout}\\
\bar{\rho}_{i+1\mid k}
&=
\lambda_\rho\bar{\rho}_{i\mid k}
+(1-\lambda_\rho)p_{\zeta_1}.
\label{eq:predictive_unlearnability_recursion}
\end{align}
\end{subequations}
where
$i\mid k$ denotes the $i$th prediction step conditioned on the information
available at transition $k$.
Along the prediction horizon, the residual input is constructed as
$\bm{s}_{i\mid k}=
[\bm{x}_{i\mid k}^{\top}\ \bm{o}_{i\mid k}^{\top}\
\bm{u}_{i\mid k}^{\top}]^{\top}$.
Throughout the rollout, $\bm{o}_{i\mid k}=\bm{o}_k$ is held frozen because
future policy observations are unavailable.
Similarly, future residuals have not yet been observed, so the
posterior cannot be updated at each prediction step.
The predictive probability is therefore initialized from the latest
posterior, $\bar{\rho}_{0\mid k}=\rho^{+}$.
Using the prior probability of the unlearnability condition,
$p_{\zeta_1}$, as the long-term prior yields the predictive recursion in
\eqref{eq:predictive_unlearnability_recursion}, where $\lambda_\rho\in(0,1)$
is a persistence factor controlling how much of the current probability
is retained at each prediction step. In the absence of new residual
observations, $\bar{\rho}_{i\mid k}$ converges to the prior probability
$p_{\zeta_1}$ as the prediction horizon extends.
To support optimization of the resulting rollout, the differentiability
of the RFF feature map in \eqref{eq:rff_map} allows the derivative of the
feature matrix to be computed analytically as
\begin{equation}
\frac{\mathrm{d}\bm{H}(\bm{s})}{\mathrm{d}\bm{s}}
=
-\sqrt{\frac{2}{D}}\,\bm{I}_{n_d}\otimes
\left[
\bm{\Omega}^{\top}
\operatorname{diag}\!\left(\sin(\bm{\Omega}\bm{s}+\bm{b})\right)
\right],
\label{eq:rff_design_matrix_derivative}
\end{equation}
where $\bm{\Omega}:=[\bm{\omega}_1\ \cdots\ \bm{\omega}_D]^\top
\in\mathbb{R}^{D\times n_s}$, and
$\bm{b}:=[b_1\ \cdots\ b_D]^\top\in\mathbb{R}^{D}$.
The analytic derivative can be used in the optimization of the MPC
decision variables $\bm{u}_{i\mid k}$ and the corresponding candidate
states $\bm{x}_{i\mid k}$ through the rollout.

With the learned residual model rolled out over the prediction horizon,
the robot's dynamics, cost, and constraints can be
designed to account for uncertainty during online learning and model
underfitting in nonreference operating conditions. 
Under reference operating conditions, the dynamics is designed to use the learned residual
mean:
\begin{equation}
\bm{f}_{\mathrm{L},i\mid k}
:=
\bm{f}_{\mathrm{nom}}(\bm{x}_{i\mid k},\bm{u}_{i\mid k})
+\bm{B}_d\eta_{i\mid k}\hat{\bm{\mu}}_{i\mid k}.
\label{eq:uncertainty_regulated_learnable_dynamics}
\end{equation}
The factor $\eta_{i\mid k}\in[0,1]$ determines how much residual
compensation is used based on learning uncertainty and is chosen as
\begin{equation}
\eta_{i\mid k}
:=
\left[1-\frac{r_{i\mid k}}{m_{i\mid k}}\right]_+,
\label{eq:uncertainty_adaptive_residual_gain}
\end{equation}
where $[z]_+:=\max\{z,0\}$. The uncertainty radius $r_{i\mid k}$ and
the predicted residual magnitude $m_{i\mid k}$ are defined as
\begin{subequations}
\label{eq:mpc_residual_confidence_measures}
\begin{align}
r_{i\mid k}
&:=
\beta_k\sqrt{\lambda_{\max}\!\left(
\bm{\Sigma}_0^{-1/2}\bm{A}_{i\mid k}\bm{\Sigma}_0^{-1/2}
\right)},
\label{eq:mpc_residual_uncertainty_radius}\\
m_{i\mid k}
&:=
\sqrt{\hat{\bm{\mu}}_{i\mid k}^{\top}
\bm{\Sigma}_0^{-1}\hat{\bm{\mu}}_{i\mid k}}.
\label{eq:mpc_residual_mean_magnitude}
\end{align}
\end{subequations}
Here, $\beta_k>0$ is a prescribed uncertainty scaling factor, and
$\lambda_{\max}(\cdot)$ denotes the largest eigenvalue.
The reference residual covariance $\bm{\Sigma}_0$ is used to normalize
both quantities. Thus, $\eta_{i\mid k}$ accounts for learning uncertainty
when incorporating the learned residual into the prediction dynamics,
helping prevent unreliable residual compensation from amplifying control
errors.

\begin{lemma}[\mdseries Rollout Prediction Error Bound]
\label{lem:residual_compensation_error}
Under the conditions of Theorem~\ref{thm:reference_mean_accuracy}, for the
dynamics in \eqref{eq:uncertainty_regulated_learnable_dynamics} with
$\eta_{i\mid k}$ in \eqref{eq:uncertainty_adaptive_residual_gain}, the
compensation error
$\bm e^{\mathrm{comp}}_{i\mid k}:=\bm\mu(\bm s_{i\mid k})
-\eta_{i\mid k}\hat{\bm\mu}_{i\mid k}$
satisfies the following inequality at every prediction step along the rollout
$\bm s_{i\mid k}\in\mathcal D$,
\begin{equation}
\|\bm e^{\mathrm{comp}}_{i\mid k}\|_{\bm\Sigma_0^{-1}}^2
\le
\|\bm\mu(\bm s_{i\mid k})\|_{\bm\Sigma_0^{-1}}^2
-
[m_{i\mid k}-r_{i\mid k}]_+^2.
\label{eq:residual_compensation_nondegradation}
\end{equation}
Here, $r_{i\mid k}$ is the uncertainty radius in
\eqref{eq:mpc_residual_uncertainty_radius}.
\end{lemma}

Lemma~\ref{lem:residual_compensation_error} shows that the factor
$\eta_{i\mid k}$ ensures that the prediction error at each rollout prediction
step does not exceed that of the nominal model. This eliminates the adverse
effects of online learning errors and uncertainty on prediction accuracy.

The cost in reference operating conditions combines end-effector
tracking and input regularization. Using
the end-effector tracking error $\bm{e}_{e,i\mid k}$ according to \eqref{eq:end_effector_tracking_error},
the corresponding cost is
\begin{equation}
\ell_{\mathrm{L},i\mid k}
:=
\|\bm{e}_{e,i\mid k}\|_{\bm{Q}_e}^{2}
+\|\bm{u}_{i\mid k}\|_{\bm{R}}^{2},
\label{eq:learnable_stage_cost}
\end{equation}
where the weighting matrices $\bm{Q}_e$, $\bm{Q}_x$ are positive semidefinite, and $\bm{R}$ 
is positive definite.

Under nonreference operating conditions, 
the base prediction is anchored to its measured state, while the
manipulator kinematics are retained as
\begin{equation}
\bm{f}_{\mathrm{con},i\mid k}
:=
\begin{bmatrix}
\bm{x}_{b,k}\\
\bm{q}_{a,i\mid k}+\Delta t\bm{u}_{a,i\mid k}
\end{bmatrix}.
\label{eq:conservative_surrogate_transition}
\end{equation}
This conservative surrogate removes uncertain predicted base motion from
the rollout, allowing the frozen locomotion policy to 
regulate base motion within its own stable operating region and preserve
robot stability. For the manipulator, the control objective is
designed to keep the manipulator in a safe configuration. The cost is
\begin{equation}
\ell_{\mathrm{U},i\mid k}
:=
\|\bm{q}_{a,i\mid k}-\bm{q}_{a}^{\mathrm{safe}}\|_{
\bm{Q}_{a}^{\mathrm{U}}}^{2}
+\|\bm{u}_{i\mid k}\|_{\bm{R}^{\mathrm{U}}}^{2}.
\label{eq:unlearnable_stage_cost}
\end{equation}
Here, $\bm{q}_{a}^{\mathrm{safe}}$ is the safe manipulator
configuration. The weighting matrix 
$\bm{Q}_{a}^{\mathrm{U}}$ is positive semidefinite, whereas 
$\bm{R}^{\mathrm{U}}$ is positive definite.

The two conditions are combined using the predictive unlearnability
probability $\bar{\rho}_{i\mid k}$. The dynamics is
\begin{equation}
\bm{f}_{\mathrm{pred},i\mid k}
:=
(1-\bar{\rho}_{i\mid k})\bm{f}_{\mathrm{L},i\mid k}
+\bar{\rho}_{i\mid k}\bm{f}_{\mathrm{con},i\mid k},
\label{eq:posterior_blended_dynamics}
\end{equation}
and the cost is
\begin{equation}
\ell_{\bar{\rho},i\mid k}
:=
(1-\bar{\rho}_{i\mid k})\ell_{\mathrm{L},i\mid k}
+\bar{\rho}_{i\mid k}\ell_{\mathrm{U},i\mid k}.
\label{eq:posterior_blended_stage_cost}
\end{equation}
In the reference condition, the controller uses the online residual
model to reduce end-effector tracking errors, with
$\eta_{i\mid k}$ regulating the correction according to learning
uncertainty. As the predictive unlearnability probability
$\bar{\rho}_{i\mid k}$ increases, the
dynamics and objective progressively favor the conservative surrogate,
and drive the manipulator towards a safe configuration. This conservative control behavior is
intended to improve robustness.

With the dynamics and cost defined in \eqref{eq:posterior_blended_dynamics} and
\eqref{eq:posterior_blended_stage_cost}, respectively, the proposed
residual-augmented MPC is formulated as
\begin{equation}
\begin{aligned}
\bm{U}_k^\star
=\arg\min_{\bm{U}_k}\quad
&\sum_{i=0}^{N_p-1}\ell_{\bar{\rho},i\mid k}\\
\mathrm{s.t.}\quad
&\bm{x}_{0\mid k}=\bm{x}_k,\\
&\bar{\rho}_{0\mid k}=\rho^{+},\\
&\bm{x}_{i+1\mid k}=\bm{f}_{\mathrm{pred},i\mid k},\\
&\hat{\bm{\mu}}_{i\mid k}
=\bm{H}(\bm{s}_{i\mid k})\bm{\theta}_k,\\
&\bm{A}_{i\mid k}
=\bm{H}(\bm{s}_{i\mid k})\bm{P}_k
\bm{H}(\bm{s}_{i\mid k})^\top,\\
&\bar{\rho}_{i+1\mid k}
=\lambda_\rho\bar{\rho}_{i\mid k}+(1-\lambda_\rho)p_{\zeta_1},\\
&\bm{u}_{a,i\mid k}\in\mathcal{U}_a,
\quad \bm{c}_{\ell,i\mid k}\in\mathcal{C}_{\ell},\\
&\bm{q}_{a,i\mid k}\in\mathcal{Q}_a,
\quad i=0,\ldots,N_p-1.
\end{aligned}
\label{eq:residual_augmented_mpc}
\end{equation}
The sets $\mathcal{Q}_a$, $\mathcal{U}_a$, and $\mathcal{C}_{\ell}$ impose the
manipulator joint-position, manipulator joint-velocity, and locomotion command
limits, respectively. At each transition, \eqref{eq:residual_augmented_mpc} is
solved online using the learned residual model, and only the first command
$\bm{u}_{0\mid k}^{\star}$ is applied.


\section{Simulation Experiments}\label{sec:sim_experiments}
Simulation experiments evaluate tracking accuracy, disturbance recovery,
and adaptation across whole-body configurations.\footnote{\label{fn:experiment_video}Simulation and hardware experiment videos are available at
\url{https://youtu.be/H-u8buk86C0}.}
Three methods are compared:
\begin{enumerate}
\item \textbf{Nominal MPC}: The reduced-order formulation in
\cite{ma2022combining,li2022bridging}, using
$\bm f_{\mathrm{nom}}$ in \eqref{eq:nominal_wholebody}
without residual learning or compensation.
\item \textbf{SSI-MPC}: The online residual-learning formulation in
\cite{zhou2025simultaneous}, which adopts a homogeneous RFF residual model
and estimates its parameters online using gradient descent.
\item \textbf{Ours}: The proposed method combines online residual learning
in Algorithm~\ref{alg:online_residual_learning} with residual-augmented
MPC in \eqref{eq:residual_augmented_mpc}, as shown in Fig.~\ref{fig:method_framework}.
\end{enumerate}
All methods retain the same frozen \textit{DreamWaQ} locomotion policy
\cite{nahrendra2023dreamwaq}, follow the same reference, and use the OCS2
SQP solver \cite{ocs2}. The prediction horizon and discretization interval
are $1.0$~s and $0.02$~s. The shared nominal parameters are
$z^d=0.50$~m, $\tau_z=0.51$~s, $\tau_\theta=0.67$~s, and
$\tau_\phi=0.51$~s. Both learners use $D=64$ RFFs, kernel standard
deviation $0.25$, and random seed $44$.
Simulations use an M1 quadruped \cite{zhang2026vra} with a manipulator
in MuJoCo \cite{todorov2012mujoco}, running on an Intel Core i9-14900K
workstation. Physics is integrated at $500$~Hz. The policy, MPC, and
online learner run at $50$~Hz.

\begin{figure*}[!t]
\centering
{\footnotesize
\definecolor{simDesired}{HTML}{222222}
\definecolor{simNominal}{HTML}{888888}
\definecolor{simOurs}{HTML}{D62728}
\setlength{\tabcolsep}{0.85em}
\begin{tabular}{@{}cccc@{}}
\raisebox{0.35ex}{\tikz{\draw[simDesired,line width=1.05pt,dash pattern=on 3pt off 2pt] (0,0) -- (1.75em,0);}}\,Desired &
\simlegendline{simNominal}{1.0pt}\,Nominal MPC &
\simlegendline{simSSI}{1.0pt}\,SSI-MPC &
\simlegendline{simOurs}{1.0pt}\,Ours
\end{tabular}\par}
\vspace{0.35em}
\subfloat[Helical trajectory tracking in simulation.\label{fig:sim_results_snapshot}]{%
\includegraphics[width=0.32\textwidth,height=0.29\textwidth,keepaspectratio]{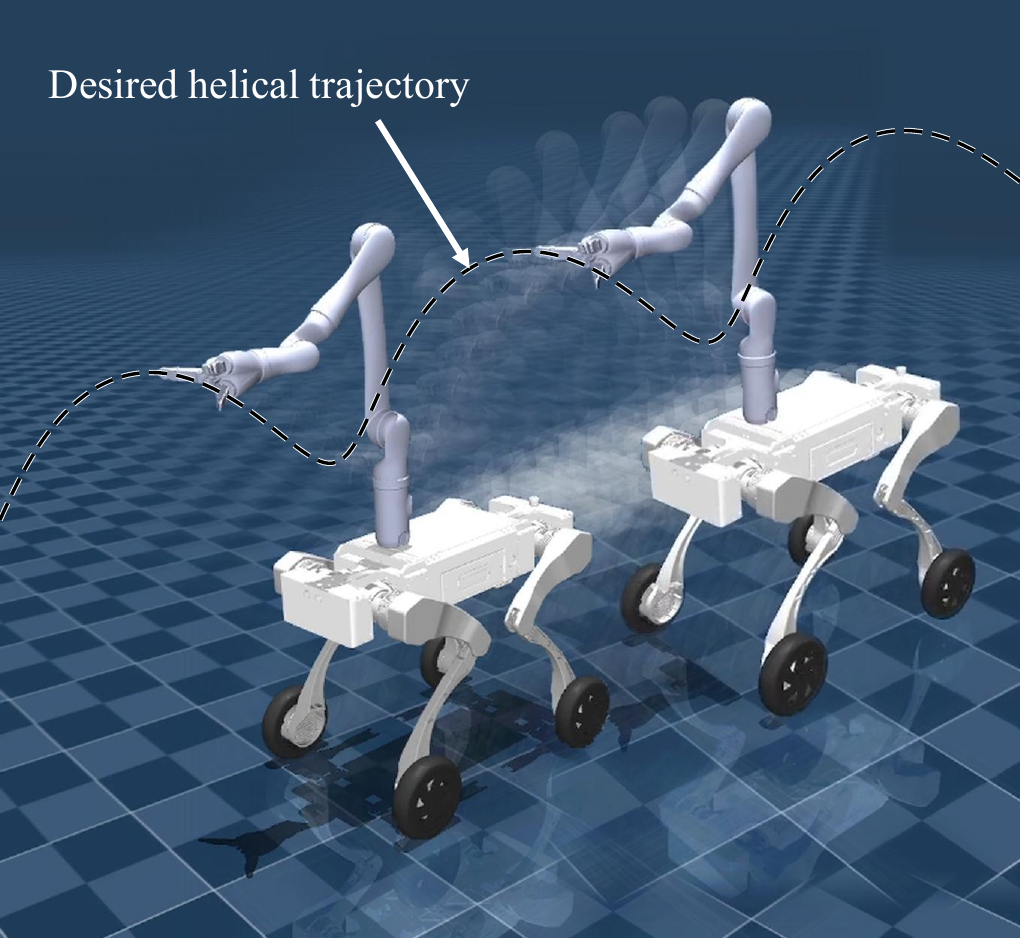}}
\hfill
\subfloat[Cartesian end-effector trajectories.\label{fig:sim_results_3d}]{%
\includegraphics[width=0.32\textwidth,height=0.29\textwidth,keepaspectratio]{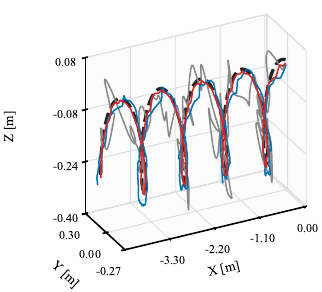}}
\hfill
\subfloat[$Y$--$Z$ projection of one complete revolution.\label{fig:sim_results_yz}]{%
\includegraphics[width=0.32\textwidth,height=0.29\textwidth,keepaspectratio]{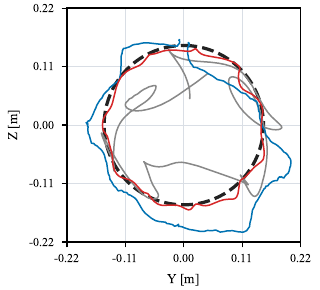}}
\caption{Simulation results of end-effector tracking of a helical trajectory. 
The black dashed curve denotes the desired
trajectory, while the gray, blue, and red curves correspond to
\textbf{Nominal MPC}, \textbf{SSI-MPC}, and the proposed method \textbf{Ours}, respectively. 
(a) Successive robot configurations during helical trajectory tracking
in simulation.
(b) Cartesian end-effector trajectories.
(c) $Y$--$Z$ projection of one complete revolution.
The proposed method \textbf{Ours} more closely follows the desired trajectory
than \textbf{Nominal MPC} and \textbf{SSI-MPC}.}
\label{fig:sim_tracking_result_ee_error}
\par\vspace{1em}
\subfloat[End-effector position tracking error.\label{fig:sim_results_error}]{%
\includegraphics[width=0.32\textwidth,height=0.29\textwidth,keepaspectratio]{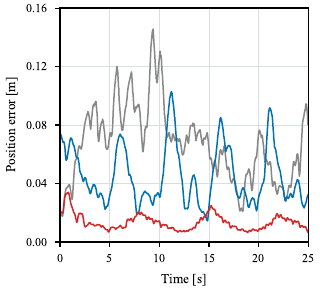}}
\hfill
\subfloat[Residual prediction error norm.\label{fig:sim_results_residual_norm}]{%
\includegraphics[width=0.32\textwidth,height=0.29\textwidth,keepaspectratio]{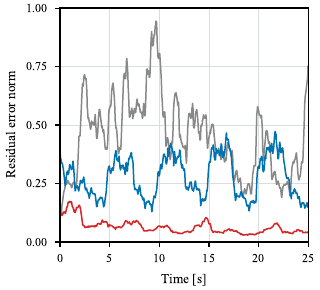}}
\hfill
\subfloat[Multi-step residual prediction skill.\label{fig:sim_results_mpc_prediction}]{%
\includegraphics[width=0.32\textwidth,height=0.29\textwidth,keepaspectratio]{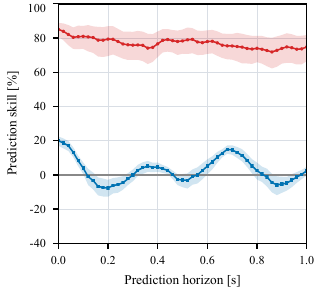}}
\caption{Simulation results of end-effector helical trajectory tracking.
The gray, blue, and red curves correspond to
\textbf{Nominal MPC}, \textbf{SSI-MPC}, and the proposed method \textbf{Ours}, respectively. 
\textbf{Nominal MPC} uses a zero-residual prediction.
(a) End-effector position tracking error.
(b) Residual prediction error norm.
(c) Multi-step residual prediction skill, where the solid curves denote the mean skill,
and the shaded regions denote the pointwise 95\% block-bootstrap confidence
intervals.
The zero-residual prediction of \textbf{Nominal MPC} gives zero skill throughout the horizon.
The proposed method \textbf{Ours} achieves smaller tracking and residual prediction errors
than \textbf{Nominal MPC} and \textbf{SSI-MPC}, while retaining the highest predictive skill
throughout the $1.0$ s MPC horizon.}
\label{fig:sim_tracking_result_residual_error}
\end{figure*}
\subsection{End-Effector Helical Trajectory Tracking}
\label{subsec:sim_precision}
The first experiment evaluates tracking of a cylindrical helix using
an M1 quadruped with a Kinova Gen2 manipulator \cite{kinova_gen2},
mounted $0.15$~m forward of the base center as shown in
Fig.~\ref{fig:sim_results_snapshot}. The desired end-effector position is
\begin{equation}
\bm p^d(t)=
\begin{bmatrix}
x_0-v_x t\\
y_c+r\cos(\omega t+\varphi_0)\\
z_c+r\sin(\omega t+\varphi_0)
\end{bmatrix},
\qquad \omega=\frac{2\pi}{T},
\label{eq:sim_helical_reference}
\end{equation}
where $r=0.15$~m, $T=5$~s, and $v_x=0.2$~m/s.

Figs.~\ref{fig:sim_results_3d} and~\ref{fig:sim_results_yz} show
pronounced trajectory distortions for \textbf{Nominal MPC}.
\textbf{SSI-MPC} improves tracking, but its projection retains an offset
and deformation. Correspondingly, Figs.~\ref{fig:sim_results_error}
and~\ref{fig:sim_results_residual_norm} show smaller errors than the
nominal baseline, together with recurrent transient peaks.
\textbf{Nominal MPC} leaves the policy-induced residual uncompensated.
\textbf{SSI-MPC} learns this mismatch through homogeneous regression,
although similar observable inputs correspond to heterogeneous latent
responses. Pooling these observations shifts the estimated mean away
from the reference response, leaving a compensation bias and steady-state
tracking offset. The additional between-condition covariance in
\eqref{eq:marginal_residual_covariance} increases finite-data update
variability, causing fluctuating compensation and transient tracking errors.
The sustained residual error and recurrent peaks in
Fig.~\ref{fig:sim_results_residual_norm} are consistent with these effects.

\textbf{Ours} more closely follows the helix and its circular projection,
with tracking errors mainly around $0.01$~m after initial adaptation.
Fig.~\ref{fig:sim_results_residual_norm} also shows a lower sustained
prediction error and smaller transient peaks. Sharing the reference mean
between the Gaussian components prevents their proportions from shifting
the predictive center. Covariance inflation and posterior weighting
limit the influence of heterogeneous deviations on updates, supporting
more consistent residual compensation.

For a prediction lead of $h$ sampling intervals, the residual prediction
skill is defined as
\begin{equation}
\mathcal S(h)=100\left(1-
\frac{\sum_k\|\bm d_{k+h}-\hat{\bm\mu}_{k+h\mid k}\|_2^2}
{\sum_k\|\bm d_{k+h}\|_2^2}\right)\%.
\label{eq:sim_prediction_skill}
\end{equation}
Here, $\bm d_{k+h}$ is the measured residual and
$\hat{\bm\mu}_{k+h\mid k}$ is its prediction formed at step $k$.
Positive skill indicates improvement over zero-residual prediction.
In Fig.~\ref{fig:sim_results_mpc_prediction}, \textbf{Nominal MPC}
has zero skill, while \textbf{SSI-MPC} repeatedly falls below zero.
\textbf{Ours} retains approximately $72\%$--$85\%$ skill, with
$75\%$ at $1.0$~s and a pointwise $95\%$ block-bootstrap interval
of $66\%$--$82\%$ within the recorded run.
Without epistemic uncertainty regulation, the SSI-MPC optimizer can
favor states and commands whose apparent benefit arises from an
underfitted residual model, propagating unreliable compensation along
the rollout. For \textbf{Ours}, the gain in
\eqref{eq:uncertainty_adaptive_residual_gain} attenuates uncertain
corrections along each candidate rollout, discouraging this behavior.
The consistently positive skill supports useful compensation throughout
the MPC horizon.

\begin{figure*}[!t]
\centering
{\footnotesize
\definecolor{simNominal}{HTML}{888888}
\definecolor{simSSI}{HTML}{0072B2}
\definecolor{simOurs}{HTML}{D62728}
\renewcommand{\simlegendline}[2]{%
  \raisebox{0.35ex}{\textcolor{#1}{\rule{1.75em}{#2}}}}
\renewcommand{\simlegendpatch}[1]{%
  \raisebox{0.05ex}{\textcolor{#1}{\rule{1.75em}{0.65em}}}}
\setlength{\tabcolsep}{0.85em}
\makebox[\textwidth][c]{\hspace*{0.0563\textwidth}%
\begin{tabular}{@{}cccc@{}}
\simlegendline{simNominal}{1.0pt}\,Nominal MPC &
\simlegendline{simSSI}{1.0pt}\,SSI-MPC &
\simlegendline{simOurs}{1.0pt}\,Ours &
\simlegendpatch{simForce}\,Force interval
\end{tabular}}\par}
\vspace{0.35em}
\subfloat[Multi-directional external disturbances in simulation\label{fig:sim_disturbance_scenarios}]{%
\raisebox{0.0145\textwidth}[\dimexpr\height+0.0145\textwidth\relax][0pt]{%
\includegraphics[width=0.49\textwidth]{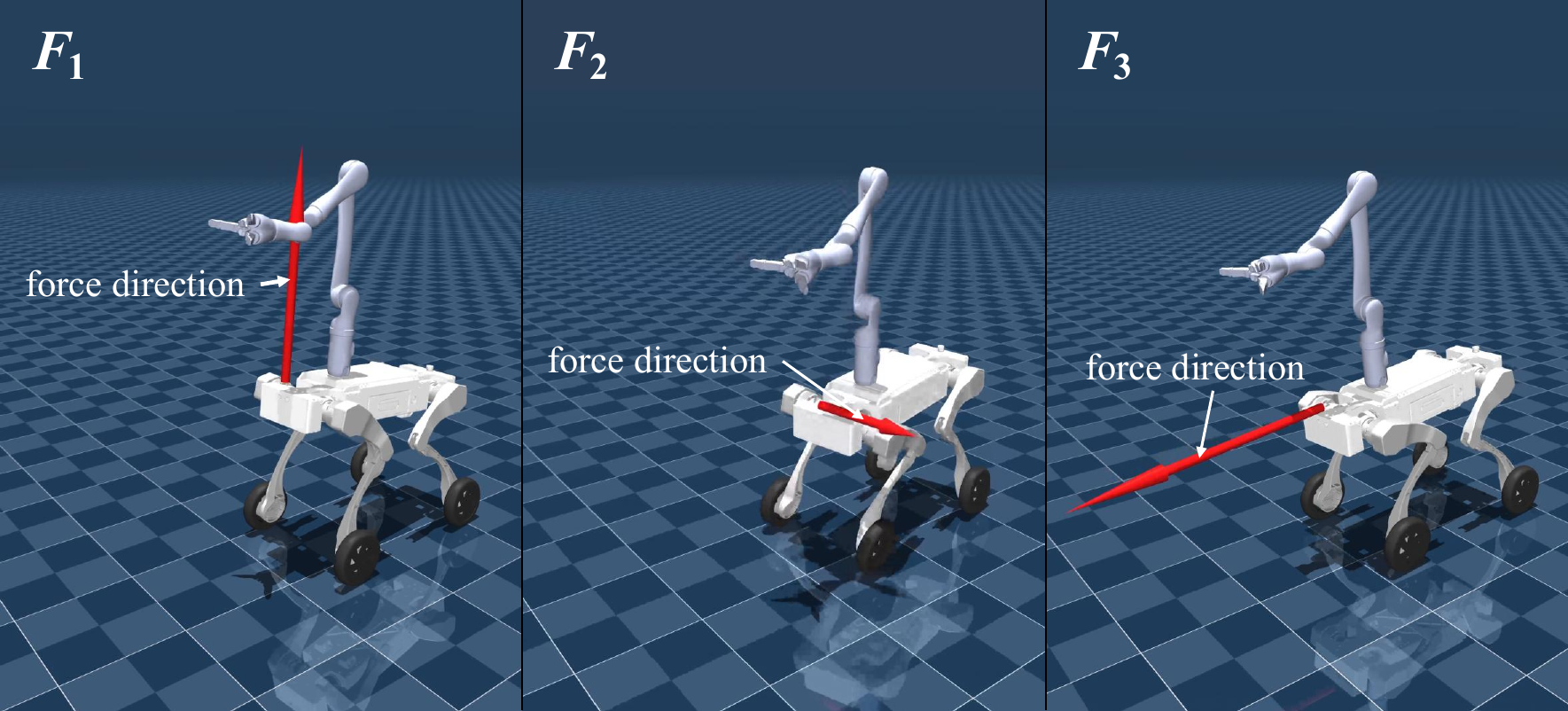}}}
\hfill
\subfloat[End-effector position tracking error.\label{fig:sim_disturbance_tracking}]{%
\includegraphics[width=0.49\textwidth]{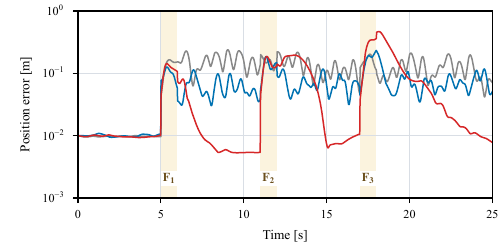}}\\[0.6em]
\subfloat[Residual prediction error norm.\label{fig:sim_disturbance_prediction}]{%
\includegraphics[width=0.49\textwidth]{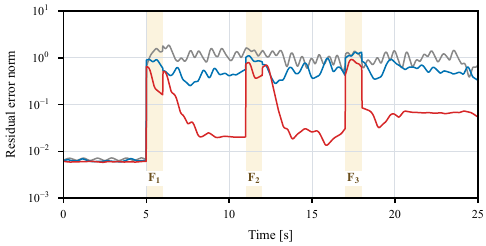}}
\hfill
\subfloat[Posterior probability.\label{fig:sim_disturbance_posterior}]{%
\includegraphics[width=0.49\textwidth]{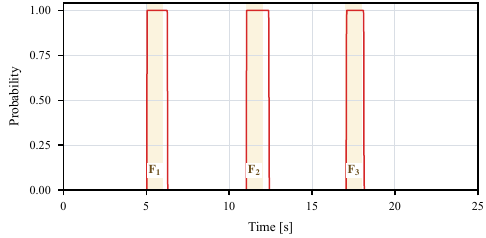}}
\caption{Simulation results of multi-directional external disturbances.
The gray, blue, and red curves correspond to
\textbf{Nominal MPC}, \textbf{SSI-MPC}, and the proposed method \textbf{Ours}, respectively. 
Yellow shaded regions indicate the force intervals $F_1$--$F_3$.
(a) Multi-directional external disturbances in simulation, with $100$~N forces sequentially
applied along the three Cartesian axes. Red arrows indicate force directions.
(b) End-effector position tracking error.
(c) Residual prediction error norm.
(d) Posterior probability $\rho_k^{+}$ defined in \eqref{eq:posterior_unlearnability_probability}.
The proposed method \textbf{Ours} identifies nonreference responses
through the posterior probability, achieving smaller tracking and
residual prediction errors after disturbance removal than either
\textbf{Nominal MPC} or \textbf{SSI-MPC}.
}
\label{fig:sim_disturbance_outcomes}
\end{figure*}
\subsection{Multi-Directional External Disturbances}
\label{subsec:sim_disturbance}
The second experiment uses strong disturbances to evaluate protection
against heterogeneous residual contamination. A fixed end-effector pose
is prescribed while $100$~N forces are sequentially applied along the
three positive Cartesian directions as shown in
Fig.~\ref{fig:sim_disturbance_scenarios}. The intervals $F_1$--$F_3$
introduce nonreference conditions that strongly excite latent responses.
The frozen policy moves with the external load to maintain balance,
temporarily departing from its task-related reference response.

Figs.~\ref{fig:sim_disturbance_tracking}
and~\ref{fig:sim_disturbance_prediction} show substantial deviations
for both baselines. After force removal, \textbf{Nominal MPC} retains
large prediction errors and tracking fluctuations. \textbf{SSI-MPC}
reduces these errors, confirming the benefit of residual learning,
but does not rapidly recover its pre-disturbance prediction accuracy.
Its homogeneous updates incorporate the forced recovery observations
into the same function used under reference conditions. The resulting
mean shift and update variability continue to affect compensation
after the latent condition returns towards the reference regime.
The persistent prediction errors and tracking offsets are consistent
with this contamination.

\textbf{Ours} recovers substantially smaller prediction and tracking
errors after the disturbances. Following $F_1$, $F_2$ and $F_3$, the position
error returns to $0.01$~m. In
Fig.~\ref{fig:sim_disturbance_posterior}, $\rho_k^+$ rises towards one
during the nonreference responses and returns towards zero afterward.
The corresponding reduction of $\gamma_k=1-\rho_k^+$ suppresses
the update gain in \eqref{eq:posterior_weighted_rls_gain}, limiting
contamination of the reference mean. Once reference conditions resume,
the preserved model supports compensation and continued adaptation.

The posterior also shifts the MPC dynamics and objective towards
conservative control through \eqref{eq:posterior_blended_dynamics}
and~\eqref{eq:posterior_blended_stage_cost}. Reduced reliance on
uncertain base predictions allows the frozen policy to prioritize
stabilization before precise tracking resumes. Thus, posterior
inference supports both protection of online learning and reuse of
the policy's existing disturbance recovery behavior.

\begin{figure}[!t]
\centering
{\footnotesize
\definecolor{simNominal}{HTML}{888888}
\definecolor{simSSI}{HTML}{0072B2}
\definecolor{simOurs}{HTML}{D62728}
\setlength{\tabcolsep}{0.7em}
\makebox[\columnwidth][c]{\hspace*{0.0563\columnwidth}%
\begin{tabular}{@{}ccc@{}}
\simlegendpatch{simNominal}\,Nominal MPC &
\simlegendpatch{simSSI}\,SSI-MPC &
\simlegendpatch{simOurs}\,Ours
\end{tabular}}\par}
\vspace{0.2em}
\captionsetup[subfloat]{format=configaxescenter,singlelinecheck=true}
\subfloat[Position tracking error.\label{fig:sim_configuration_comparison_kinova_tracking}]{%
\includegraphics[width=0.50\columnwidth]{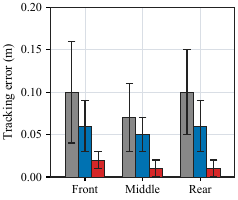}}
\hfill
\subfloat[Residual prediction error.\label{fig:sim_configuration_comparison_kinova_prediction}]{%
\includegraphics[width=0.50\columnwidth]{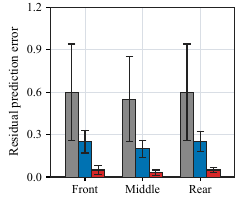}}
\caption{Simulation results of end-effector trajectory tracking with the
Kinova Gen2 \cite{kinova_gen2} manipulator at different mounting positions.
Front, middle, and rear correspond to longitudinal offsets of
$+0.15$, $0$, and $-0.15$~m from the quadruped's geometric center,
respectively.
(a) End-effector position tracking error.
(b) Residual prediction error.
Gray, blue, and red bars represent \textbf{Nominal MPC},
\textbf{SSI-MPC}, and \textbf{Ours}, respectively.
Bar heights and error bars indicate the mean and standard deviation
across repeated experiments. 
\textbf{Ours} consistently achieves lower tracking and residual
prediction errors than \textbf{Nominal MPC} and \textbf{SSI-MPC}
across all mounting positions.}
\label{fig:sim_configuration_comparison_kinova}
\vspace{0.8em}
\subfloat[Position tracking error.\label{fig:sim_configuration_comparison_piper_tracking}]{%
\includegraphics[width=0.50\columnwidth]{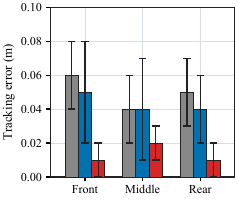}}
\hfill
\subfloat[Residual prediction error.\label{fig:sim_configuration_comparison_piper_prediction}]{%
\includegraphics[width=0.50\columnwidth]{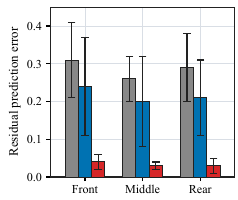}}
\caption{Simulation results of end-effector trajectory tracking with the
AgileX Piper \cite{agilex_piper_sdk} manipulator at different mounting
positions.
Front, middle, and rear correspond to longitudinal offsets of
$+0.15$, $0$, and $-0.15$~m from the quadruped's geometric center,
respectively.
(a) End-effector position tracking error.
(b) Residual prediction error.
Gray, blue, and red bars represent \textbf{Nominal MPC},
\textbf{SSI-MPC}, and \textbf{Ours}, respectively.
Bar heights and error bars indicate the mean and standard deviation
across repeated experiments.
\textbf{Ours} consistently achieves lower tracking and residual
prediction errors than \textbf{Nominal MPC} and \textbf{SSI-MPC}
across all mounting positions.}
\label{fig:sim_configuration_comparison_piper}
\end{figure}
\subsection{Cross-Embodiment Online Learning}
\label{subsec:sim_generalization}
The third experiment evaluates whether online learning accommodates
different manipulators and mounting positions with consistent accuracy.
The Kinova Gen2 \cite{kinova_gen2} and AgileX Piper
\cite{agilex_piper_sdk} are each mounted at longitudinal offsets of
$+0.15$, $0$, and $-0.15$~m from the base center, giving six
whole-body configurations. Each performs the trajectory in
\eqref{eq:sim_helical_reference} with the same frozen policy,
learning configuration, and MPC settings.

Figs.~\ref{fig:sim_configuration_comparison_kinova}
and~\ref{fig:sim_configuration_comparison_piper} report the mean
errors and their standard deviations across repeated experiments.
\textbf{Nominal MPC} has tracking errors of $0.04$--$0.10$~m
and residual prediction errors of $0.26$--$0.60$ across configurations.
\textbf{SSI-MPC} reduces these ranges to $0.04$--$0.06$~m and
$0.20$--$0.25$, respectively, showing that online compensation
partly accommodates changes in whole-body dynamics.
\textbf{Ours} maintains tracking errors of $0.01$--$0.02$~m
and residual prediction errors of $0.03$--$0.05$ for both manipulators
at all three mounting positions.

Changing the manipulator and its mounting position changes the mass
distribution and coupling seen by the frozen policy.
A fixed nominal
model therefore leaves configuration-dependent prediction errors.
Online learning adapts the residual response from the transitions
observed under each configuration. Posterior weighting additionally
limits interference from heterogeneous observations, helping retain
a consistent compensation model despite differences in latent
responses. The low errors across all six configurations support
adaptation through online residual learning without retraining the
locomotion policy for each manipulator or mounting position.

\section{Hardware Experiments}\label{sec:hardware_experiments}
Hardware experiments evaluate online adaptation across five
loco-manipulation tasks.\footref{fn:experiment_video}
The platform is an M1 wheel-legged robot \cite{zhang2026vra}
with a Kinova Gen2 manipulator \cite{kinova_gen2}
as shown in Fig.~\ref{fig:hardware_show}. The \textit{DreamWaQ} policy
\cite{nahrendra2023dreamwaq} remains frozen. The policy, MPC, and online learner run at $50$~Hz on an ASUS NUC
equipped with an Intel Core Ultra 7 155H CPU. The MPC problem is
solved using the SQP solver provided by OCS2 \cite{ocs2}.
The RFF settings and $1.0$~s
prediction horizon with $0.02$~s intervals match the simulations.
The same online learning and control formulations are used throughout.

\begin{figure*}[!t]
\centering
\makebox[\textwidth][s]{%
\subfloat[A snapshot of the trajectory track.\label{fig:hardware_tracking_results_a}]{%
\includegraphics[width=0.24\textwidth]{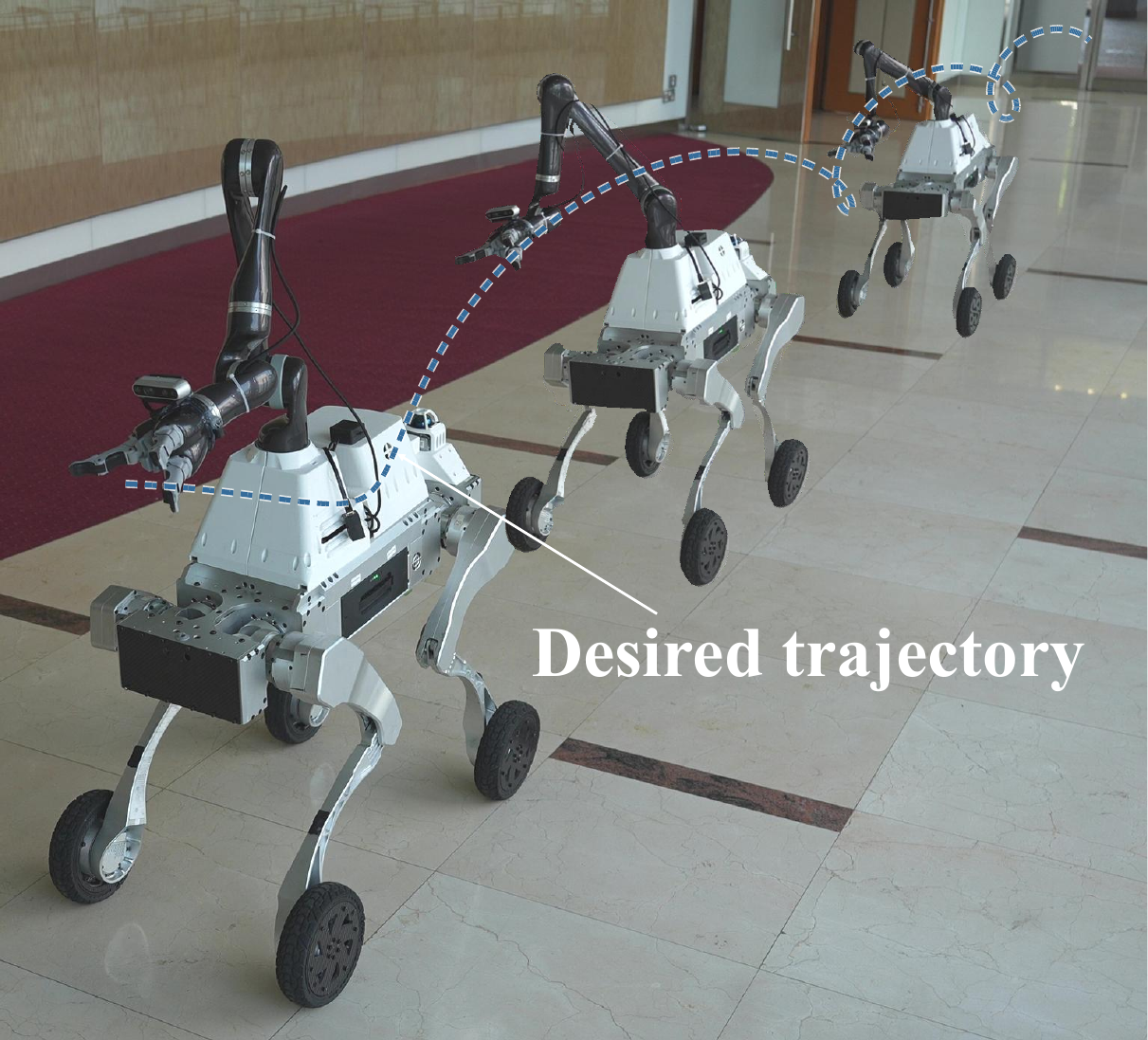}}
\hfill
\subfloat[End-effector tracking error.\label{fig:hardware_tracking_results_b}]{%
\includegraphics[width=0.24\textwidth]{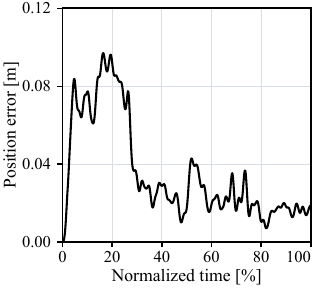}}%
\hfill
\subfloat[Residual prediction error.\label{fig:hardware_tracking_results_c}]{%
\includegraphics[width=0.24\textwidth]{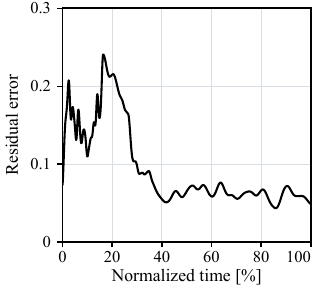}}%
\hfill
\subfloat[MPC prediction skill.\label{fig:hardware_tracking_results_d}]{%
\includegraphics[width=0.24\textwidth]{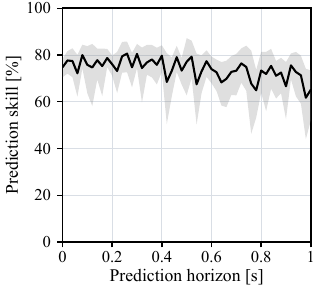}}
}%
\caption{Hardware experimental results for helical trajectory tracking.
(a) Snapshots of the trajectory tracking. The dashed
curve denotes the desired end-effector trajectory.
(b) End-effector position tracking error.
(c) One-step residual prediction error norm.
(d) Multi-step residual prediction skill, where the solid curve
denotes the mean and the shaded region indicates the pointwise
95\% confidence interval.
Time in (b) and (c) is normalized by the task duration.
The proposed method reduces the end-effector tracking and residual
prediction errors through online learning, retaining the high
prediction skill throughout the $1.0$~s MPC horizon.
}
\label{fig:hardware_tracking_results}
\par\vspace{0.8em}
\centering
\makebox[\textwidth][s]{%
\subfloat[A snapshot of the object grasping.\label{fig:hardware_grasping_results_a}]{%
\includegraphics[width=0.24\textwidth]{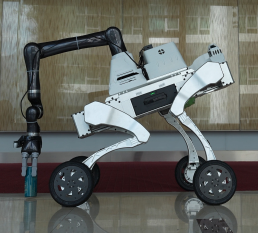}}%
\hfill
\subfloat[End-effector tracking error.\label{fig:hardware_grasping_results_b}]{%
\includegraphics[width=0.24\textwidth]{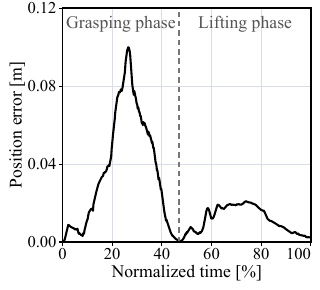}}%
\hfill
\subfloat[Residual prediction error.\label{fig:hardware_grasping_results_c}]{%
\includegraphics[width=0.24\textwidth]{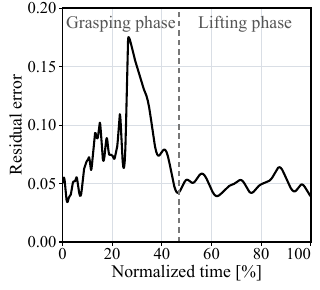}}%
\hfill
\subfloat[MPC prediction skill.\label{fig:hardware_grasping_results_d}]{%
\includegraphics[width=0.24\textwidth]{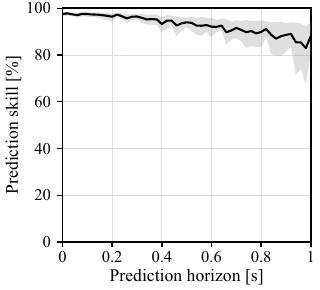}}
}%
\caption{Hardware experimental results for object grasping.
(a) A snapshot of object grasping.
(b) End-effector position tracking error.
(c) One-step residual prediction error norm.
(d) Multi-step residual prediction skill, where the solid curve
denotes the mean and the shaded region indicates the pointwise
95\% confidence interval.
Time in (b) and (c) is normalized by the task duration.
The proposed method adapts online across the grasping and lifting
phases, supporting accurate object grasp with high prediction
skill throughout the $1.0$~s MPC horizon.}
\label{fig:hardware_grasping_results}
\par\vspace{0.8em}
\centering
\makebox[\textwidth][s]{%
\subfloat[A snapshot of grasping on the move.\label{fig:hardware_moving_results_a}]{%
\includegraphics[width=0.48\textwidth]{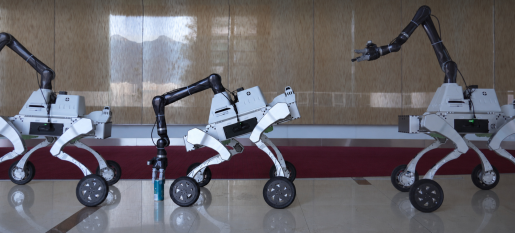}}%
\hfill
\subfloat[Tracking and residual errors.\label{fig:hardware_moving_results_b}]{%
\includegraphics[width=0.24\textwidth]{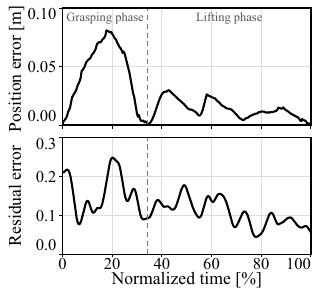}}%
\hfill
\subfloat[MPC prediction skill.\label{fig:hardware_moving_results_c}]{%
\includegraphics[width=0.24\textwidth]{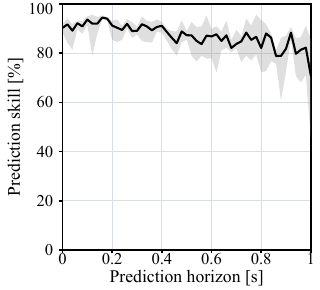}}%
}%
\caption{Hardware experimental results for grasping on the move.
(a) Snapshots of grasping on the move.
(b) End-effector position tracking error (top) and one-step residual
prediction error norm (bottom).
(c) Multi-step residual prediction skill, where the solid curve
denotes the mean and the shaded region indicates the pointwise
95\% confidence interval.
Time in (b) is normalized by the task duration.
The proposed method supports simultaneous locomotion and manipulation,
recovering low tracking errors after task-phase transitions while
maintaining high prediction skill throughout the $1.0$~s MPC horizon.}
\label{fig:hardware_moving_results}
\end{figure*}

\subsection{Trajectory Tracking}\label{subsec:hardware_tracking}
The first task examines the online learning process during helical
end-effector tracking as shown in Fig.~\ref{fig:hardware_tracking_results_a}.
The reference in \eqref{eq:sim_helical_reference} uses $r=0.20$~m,
$T=8.0$~s, and $v_x=0.40$~m/s, combining three revolutions with
floating-base translation.

Figs.~\ref{fig:hardware_tracking_results_b}
and~\ref{fig:hardware_tracking_results_c} show pronounced initial
errors, reaching approximately $0.10$~m and $0.24$, respectively.
Both decrease over the subsequent adaptation interval, with smaller
fluctuations during repeated motion. Over $30\%$--$100\%$
normalized time, the mean tracking and residual prediction errors
are approximately $0.02$~m and $0.06$.
The decrease is gradual because reliable compensation requires
informative transitions from the visited state regions. As these
observations accumulate, the learner refines the residual mean and
reduces estimation uncertainty, improving the prediction used for
command selection.
Fig.~\ref{fig:hardware_tracking_results_d} reports a
horizon-averaged skill of $74\%$, remaining positive throughout
the $1.0$~s rollout.
These results show that online learning rapidly reduces prediction
and tracking errors, enabling accurate loco-manipulation tracking
on hardware.

\subsection{Object Grasping}\label{subsec:hardware_grasping}
The second task evaluates online learning across different motion phases,
as shown in Fig.~\ref{fig:hardware_grasping_results_a}.
During the grasping phase, the manipulator reaches towards the object
and closes the gripper. During the lifting phase, it raises the grasped
object. An Intel RealSense D435 camera \cite{intel_realsense_d435}
localizes the object.

In Figs.~\ref{fig:hardware_grasping_results_b}
and~\ref{fig:hardware_grasping_results_c}, the grasping phase produces
tracking and prediction peaks of approximately $0.10$~m and $0.18$
near $27\%$ normalized time. Near the phase boundary marked at
$47\%$, these errors decrease below $0.01$~m and to approximately
$0.04$, respectively. During lifting, the tracking error exhibits
a smaller peak of $0.02$~m near $74\%$, while the residual prediction
error fluctuates around $0.05$. By task completion, the tracking
and prediction errors decrease to $0.01$~m and
$0.04$. Fig.~\ref{fig:hardware_grasping_results_d} shows a
horizon-averaged prediction skill of $93\%$.

The phase transition introduces partly unvisited states, requiring
continued adaptation. The RFF representation retains the weights
learned during grasping and shares their effect across inputs with
similar features, providing useful predictions during lifting
while new observations refine the model. The smaller tracking
transient and sustained low prediction error during lifting are
consistent with information carrying across phases, supporting
accurate task execution without restarting online learning.

\subsection{Grasping on the Move}
\label{subsec:hardware_moving_manipulation}
The third task combines locomotion, reaching, grasping, and lifting,
as shown in Fig.~\ref{fig:hardware_moving_results_a}, evaluating
adaptation between different whole-body motion states.
In Fig.~\ref{fig:hardware_moving_results_b}, the grasping phase
produces tracking and residual prediction peaks of approximately
$0.08$~m near $18\%$ normalized time and $0.25$ near $20\%$,
respectively. Near the marked phase boundary at approximately
$34\%$, these errors decrease below $0.01$~m and to approximately
$0.09$. During lifting, smaller tracking and prediction peaks
reach $0.03$~m near $43\%$ and $0.18$ near $49\%$, respectively,
followed by an overall decrease with continued fluctuations.
By task completion, tracking error is below $0.01$~m and residual
prediction error is approximately $0.06$.
Fig.~\ref{fig:hardware_moving_results_c} reports a
horizon-averaged skill of $87\%$.

The contaminated Gaussian model does not associate the reference
condition with a single fixed motion state. Reference conditions
comprise a set of routine latent states in
Definition~\ref{def:reference_operating_conditions}, and the shared
mean $\bm\mu(\bm s)$ varies with the observable input.
Transitions between locomotion and manipulation states can therefore
remain informative for learning. Posterior inference evaluates
deviations relative to this conditional response while accounting
for epistemic uncertainty, so a motion-state change alone does not
require exclusion of its observations. The renewed error reduction
and retained prediction skill are consistent with adaptation across
these routine transitions within the same model.

\subsection{Force Recovery}\label{subsec:hardware_recovery}
\begin{figure*}[!t]
\centering
\newlength{\hardwareRecoveryRowHeight}
\settoheight{\hardwareRecoveryRowHeight}{%
\includegraphics[width=0.24\textwidth]{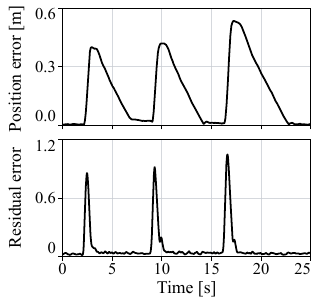}}%
\makebox[\textwidth][s]{%
\subfloat[A snapshot of the force recovery.\label{fig:hardware_result_single_a}]{%
\raisebox{\dimexpr\hardwareRecoveryRowHeight-\height\relax}[\hardwareRecoveryRowHeight][0pt]{%
\includegraphics[width=0.48\textwidth]{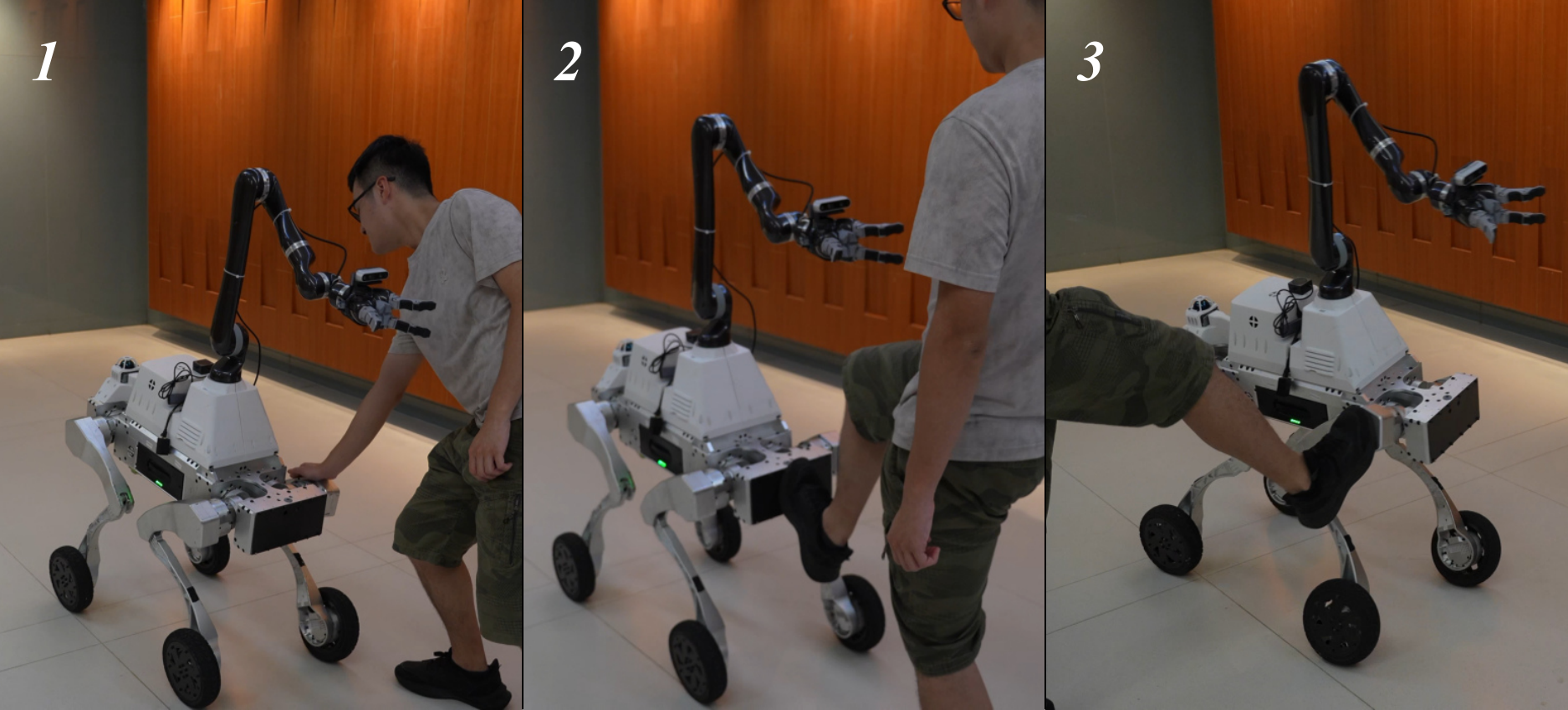}}}%
\hfill
\subfloat[Tracking and residual errors.\label{fig:hardware_result_single_b}]{%
\raisebox{\dimexpr\hardwareRecoveryRowHeight-\height\relax}[\hardwareRecoveryRowHeight][0pt]{%
\includegraphics[width=0.24\textwidth]{fig/force_recovery_errors_bc.pdf}}}%
\hfill
\subfloat[Posterior probability.\label{fig:hardware_result_single_c}]{%
\raisebox{\dimexpr\hardwareRecoveryRowHeight-\height\relax}[\hardwareRecoveryRowHeight][0pt]{%
\includegraphics[width=0.24\textwidth]{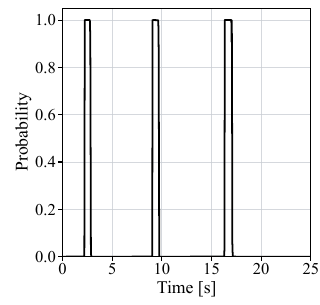}}}%
}%
\caption{Hardware experimental results for force recovery.
(a) Responses to three force perturbations along the
Cartesian axes while a fixed end-effector position is prescribed.
(b) End-effector position tracking error (top) and one-step residual
prediction error norm (bottom).
(c) Posterior unlearnability probability $\rho_k^{+}$, which rises
during each perturbation and returns close to zero afterward.
The proposed method identifies nonreference responses through the
posterior probability, supporting compliant recovery while restoring
low tracking errors after disturbance removal.}
\label{fig:hardware_result_single}
\end{figure*}

The fourth task evaluates protection of the learned model during
nonreference responses. Three manual force impulses
are applied along the positive Cartesian axes while a fixed
end-effector position is prescribed as shown in
Fig.~\ref{fig:hardware_result_single_a}. The frozen policy
responds to the perturbations by prioritizing stabilization.

Fig.~\ref{fig:hardware_result_single_b} shows tracking peaks
of approximately $0.40$, $0.42$, and $0.54$~m, accompanied by
brief residual prediction peaks. The prediction errors rapidly
subside, while the end effector returns towards the desired point
over the subsequent recovery intervals. The different durations
reflect the distinction between restoring predictive accuracy
and completing the physical return motion.
In Fig.~\ref{fig:hardware_result_single_c}, $\rho_k^+$ rises
towards one during each perturbation and returns close to zero
afterward. The resulting reduction of $\gamma_k=1-\rho_k^+$
suppresses updates from nonreference observations through
\eqref{eq:posterior_weighted_rls_gain}. The repeated recovery
without a sustained prediction-error increase is consistent with
preserving the reference mean against heterogeneous contamination.

The posterior also favors conservative MPC dynamics and objectives
through \eqref{eq:posterior_blended_dynamics}
and~\eqref{eq:posterior_blended_stage_cost}. The measured base
state anchors the rollout while the manipulator is regulated towards its
safe configuration, allowing the policy to accommodate the force.
The temporary tracking deviations therefore accompany a compliant
recovery, after which precise tracking resumes. This behavior
supports reuse of the frozen policy's existing robustness alongside
protection of online learning.

\subsection{Door Opening}\label{subsec:hardware_door_opening}
\begin{figure*}[!t]
\centering
\newlength{\hardwareDoorRowHeight}
\settoheight{\hardwareDoorRowHeight}{%
\includegraphics[width=0.24\textwidth]{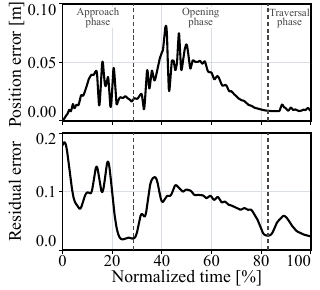}}%
\makebox[\textwidth][s]{%
\subfloat[A snapshot of the door opening.\label{fig:hardware_door_results_a}]{%
\raisebox{\dimexpr\hardwareDoorRowHeight-\height\relax}[\hardwareDoorRowHeight][0pt]{%
\includegraphics[width=0.48\textwidth]{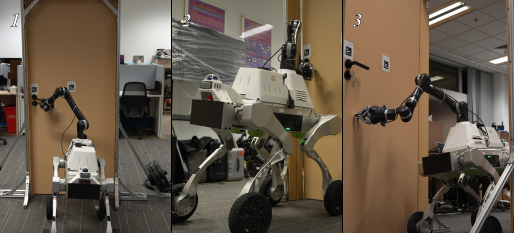}}}%
\hfill
\subfloat[Tracking and residual errors.\label{fig:hardware_door_results_b}]{%
\raisebox{\dimexpr\hardwareDoorRowHeight-\height\relax}[\hardwareDoorRowHeight][0pt]{%
\includegraphics[width=0.24\textwidth]{fig/hardware_by_task/door_opening/position_residual_errors_door_opening.pdf}}}%
\hfill
\subfloat[MPC prediction skill.\label{fig:hardware_door_results_c}]{%
\raisebox{\dimexpr\hardwareDoorRowHeight-\height\relax}[\hardwareDoorRowHeight][0pt]{%
\includegraphics[width=0.24\textwidth]{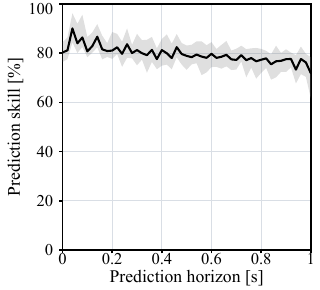}}}
}%
\caption{Hardware experimental results for door opening.
(a) Snapshots of handle approach, door opening, and traversal.
(b) End-effector position tracking error (top) and one-step residual
prediction error norm (bottom).
(c) Multi-step residual prediction skill, where the solid curve
denotes the mean and the shaded region indicates the pointwise
95\% confidence interval.
Time in (b) is normalized by the task duration.
The proposed method adapts online to sustained contact dynamics,
reducing tracking and residual prediction errors as the door
interaction is learned.}
\label{fig:hardware_door_results}
\end{figure*}

The fifth task evaluates whether online learning incorporates
environmental dynamics during sustained contact.
A $15$~kg door is used. An Intel RealSense D435 camera
\cite{intel_realsense_d435} and an ArUco marker
\cite{garrido2014aruco} localize the handle. The task comprises
approaching the handle, pushing the door open,
and traversing the doorway, as shown in
Fig.~\ref{fig:hardware_door_results_a}.

Fig.~\ref{fig:hardware_door_results_b} marks the approach-to-opening
and opening-to-traversal boundaries near $28\%$ and $82\%$
normalized time, respectively. During the approach phase, tracking
error peaks at approximately $0.05$~m near $16\%$, before tracking
and prediction errors decrease to approximately $0.02$~m and $0.02$
by the phase boundary. Contact during opening introduces renewed
transients where tracking error reaches $0.08$~m near $42\%$, while
prediction error reaches $0.12$ near $37\%$. As contact observations
accumulate, both decrease to approximately $0.01$~m and $0.025$
near the end of opening at $83\%$. During traversal, smaller
fluctuations remain, with mean tracking and prediction errors of
approximately $0.01$~m and $0.04$, respectively.
Fig.~\ref{fig:hardware_door_results_c} shows an approximately
$80\%$ horizon-averaged prediction skill, with $72\%$ at $1.0$~s.

The door resistance introduces a latent environmental contribution
transmitted through the manipulator to the base. Sustained contact
can produce a repeatable conditional response that becomes
predictable as the model adapts. It need not introduce the frequent
switches among disparate responses associated with large
between-condition dispersion. The decreasing opening-phase errors
are consistent with retaining these transitions for learning,
rather than excluding environmental interaction solely because
an external force is present. Together with force recovery, this
result illustrates the distinction between learning a repeatable
contact contribution and suppressing heterogeneous transient
responses that would contaminate the reference model.

The hardware results highlight heterogeneity as an inherent feature of
the partially observed locomotion dynamics under the frozen policy.
Accounting for the associated latent-conditioned responses enables online
learning to adapt the compensation model across different tasks and
motion phases. The resulting adaptation maintains consistent tracking
accuracy across the evaluated conditions despite changes in coordinated
base and manipulator motion. Force recovery and door opening further
demonstrate the ability to distinguish heterogeneous transient responses
from repeatable contact contributions that remain useful for learning.
Results support generalization across task, robot, and environmental dynamics.

\section{CONCLUSION}\label{sec:conclusion}
This paper proposed an online learning method for precise loco-manipulation
that compensates for heterogeneous locomotion dynamics while retaining a
pretrained locomotion policy without retraining.
The contaminated Gaussian model anchors both components to a task-relevant
reference residual mean and represents additional between-condition
dispersion through covariance inflation, reducing heterogeneity interference
in mean estimation.
An online learner based on a Gaussian process constructed with RFFs updates
the reference mean using posterior learnability weights, while residual augmented
MPC rolls out the residual mean, epistemic covariance, and regime
probability to regulate feedforward compensation and favor conservative
control under nonreference conditions.
The analysis establishes a finite-sample convergence bound for the reference
mean estimate and a rollout prediction error bound for residual compensation
over the MPC horizon.
Simulations across different manipulator configurations and hardware
experiments across diverse loco-manipulation tasks support precise
end-effector tracking, accurate one-step residual prediction, and useful
multi-step residual prediction throughout the MPC horizon.
Across manipulator configurations, the method maintains tracking
and residual prediction errors of $0.01$~m and $0.03$,
respectively, while the evaluated hardware tasks maintain consistent
end-effector tracking performance with typical errors of approximately
$0.01$~m.

The current contaminated Gaussian model provides only an approximation of heterogeneity 
and does not explicitly address how to maintain
tracking accuracy under nonreference latent conditions.
Future work will explore the use of multiple Gaussian processes to learn
the mapping from observable states to latent conditions and develop
closed-loop stability guarantees for the interaction between online
learning, MPC, and the frozen policy.

\clearpage
\appendix\label{app:posterior_discriminability}
\begin{IEEEproof}[Proof of Proposition~\ref{prop:posterior_unlearnability}]
From \eqref{eq:predictive_regime_covariances}, the predictive innovation densities are
\begin{equation}
\begin{aligned}
L_{j,k}&:=p(\bm{e}_k\mid\zeta_k=j,\mathcal{D}_k)\\
&=\mathcal{N}\!\left(\bm{e}_k,\bm{0},\bm{S}_{j,k}\right),
\quad j\in\{0,1\}.
\end{aligned}
\label{eq:innovation_regime_distribution}
\end{equation}
Bayes' rule gives the posterior odds as the prior odds multiplied by
$L_{1,k}/L_{0,k}$. Hence,
\begin{equation}
\begin{aligned}
\log\frac{\rho_k^{+}}{1-\rho_k^{+}}
&=\log\frac{p_{\zeta_1}}{1-p_{\zeta_1}}+\log\frac{L_{1,k}}{L_{0,k}}\\
&=\log\frac{p_{\zeta_1}}{1-p_{\zeta_1}}
-\frac{1}{2}\log\frac{|\bm{S}_{1,k}|}{|\bm{S}_{0,k}|}\\
&\quad+\frac{1}{2}\bm{e}_k^\top
\left(\bm{S}_{0,k}^{-1}-\bm{S}_{1,k}^{-1}\right)\bm{e}_k.
\end{aligned}
\label{eq:posterior_unlearnability_log_odds}
\end{equation}
Applying the logistic sigmoid yields
\eqref{eq:posterior_unlearnability_probability}.
\end{IEEEproof}

\begin{IEEEproof}[Proof of Lemma~\ref{thm:residual_unlearnability_discriminability}]
Define $\bm{B}^{(m)}:=\bm{\Sigma}_0^{-1/2}\bm{A}^{(m)}\bm{\Sigma}_0^{-1/2}$
and $\bm{y}^{(m)}:=\bm{\Sigma}_0^{-1/2}\bm{e}^{(m)}$.
The change from the prior log-odds is
\begin{equation}
\begin{aligned}
L^{(m)}&:=\log\frac{\rho^{+,(m)}}{1-\rho^{+,(m)}}
-\log\frac{p_{\zeta_1}}{1-p_{\zeta_1}}\\
&=-\frac12\log\frac{|\bm{B}^{(m)}+\kappa\bm{I}|}{|\bm{B}^{(m)}+\bm{I}|}
+q^{(m)},
\end{aligned}
\label{eq:discriminability_log_odds_shift}
\end{equation}
where $q^{(m)}:=\frac12(\bm{y}^{(m)})^\top
[(\bm{B}^{(m)}+\bm{I})^{-1}-(\bm{B}^{(m)}+\kappa\bm{I})^{-1}]\bm{y}^{(m)}$.
The inverse difference has eigenvalues $(\kappa-1)/[(a+1)(a+\kappa)]$
for eigenvalues $a$ of $\bm{B}^{(m)}$.
For property~(i), uniform boundedness of $\bm{A}^{(m)}$ gives
$0\le a\le\bar a<\infty$. Setting
$c_{\bar a}:=(\kappa-1)/[2(\bar a+1)(\bar a+\kappa)]>0$ yields
\begin{subequations}
\label{eq:discriminability_log_odds_divergence}
\begin{align}
\log\frac{|\bm{B}^{(m)}+\kappa\bm{I}|}{|\bm{B}^{(m)}+\bm{I}|}
&\le n_d\log\kappa,
\label{eq:discriminability_log_determinant_uniform_bound}\\
q^{(m)}&\ge c_{\bar a}\|\bm{y}^{(m)}\|_2^2,
\label{eq:discriminability_quadratic_lower_bound}\\
L^{(m)}&\ge-\frac{n_d}{2}\log\kappa+c_{\bar a}\|\bm{y}^{(m)}\|_2^2.
\label{eq:discriminability_log_odds_lower_bound}
\end{align}
\end{subequations}
Thus, $L^{(m)}\to\infty$ as $\|\bm{y}^{(m)}\|_2\to\infty$,
proving \eqref{eq:posterior_innovation_ray_limit}.
For property~(ii), let
$a_{\min}^{(m)}:=\lambda_{\min}(\bm{B}^{(m)})\to\infty$ and assume
$(\bm{e}^{(m)})^\top(\bm{S}_0^{(m)})^{-1}\bm{e}^{(m)}\le C<\infty$
uniformly in $m$. Using $\log(1+x)\le x$ and the same eigenvalue bounds gives
\begin{subequations}
\label{eq:discriminability_log_odds_vanishing_bound}
\begin{align}
0\le\log\frac{|\bm{B}^{(m)}+\kappa\bm{I}|}{|\bm{B}^{(m)}+\bm{I}|}
&\le\frac{n_d(\kappa-1)}{a_{\min}^{(m)}+1},
\label{eq:discriminability_log_determinant_vanishing_bound}\\
0\le q^{(m)}&\le\frac{(\kappa-1)C}{2(a_{\min}^{(m)}+\kappa)},
\label{eq:discriminability_quadratic_vanishing_bound}\\
|L^{(m)}|&\le\frac{(\kappa-1)(n_d+C)}{2(a_{\min}^{(m)}+1)}\to0.
\label{eq:discriminability_log_odds_vanishing_limit}
\end{align}
\end{subequations}
Continuity of the logistic sigmoid yields
\eqref{eq:posterior_epistemic_prior_limit}.
\end{IEEEproof}

\begin{IEEEproof}[Proof of
Lemma~\ref{lemma:bounded_redescending_influence}]
Let
$\bm{Q}_k:=\bm{S}_{0,k}^{-1}-\bm{S}_{1,k}^{-1}\succ\bm{0}$ and define
\begin{equation}
c_k
:=
\frac{p_{\zeta_1}}{1-p_{\zeta_1}}
\left(
\frac{|\bm{S}_{0,k}|}{|\bm{S}_{1,k}|}
\right)^{1/2}
>0.
\label{eq:posterior_odds_scale}
\end{equation}
Equations~\eqref{eq:posterior_unlearnability_log_odds} and
\eqref{eq:learnability_responsibility} give
\begin{equation}
\gamma_k
=
\frac{
1
}{
1+c_k\exp\!\left(
\frac{1}{2}\bm{e}_k^\top\bm{Q}_k\bm{e}_k
\right)
}.
\label{eq:posterior_weight_exponential_form}
\end{equation}
Let $q_k:=\lambda_{\min}(\bm{Q}_k)>0$. Then
\begin{equation}
\gamma_k
\leq
\frac{1}{c_k}
\exp\!\left(
-\frac{q_k}{2}\|\bm{e}_k\|_2^2
\right).
\label{eq:learnability_weight_exponential_bound}
\end{equation}
Moreover,
$\bm{\Sigma}_0+\gamma_k\bm{H}_k\bm{P}_k\bm{H}_k^\top
\succeq\bm{\Sigma}_0$. Using
\eqref{eq:posterior_weighted_rls_gain} and
\eqref{eq:learnability_weight_exponential_bound} therefore gives
\begin{equation}
\begin{aligned}
\|\Delta\bm{\theta}_k\|_2
&\leq
\frac{
\|\bm{P}_k\bm{H}_k^\top\|_2
}{
c_k\lambda_{\min}(\bm{\Sigma}_0)
}
\|\bm{e}_k\|_2
\exp\!\left(
-\frac{q_k}{2}\|\bm{e}_k\|_2^2
\right)\\
&\leq
\frac{
\|\bm{P}_k\bm{H}_k^\top\|_2
}{
c_k\lambda_{\min}(\bm{\Sigma}_0)
\sqrt{q_k\mathrm{e}}
}.
\end{aligned}
\label{eq:conditional_mean_influence_bound}
\end{equation}
The last inequality follows because
$r\exp(-q_kr^2/2)\leq1/\sqrt{q_k\mathrm{e}}$ for $r\geq0$.
This establishes the finite upper bound. The same exponential term converges
to zero as $\|\bm{e}_k\|_2\rightarrow\infty$, proving the redescending
limit in \eqref{eq:redescending_parameter_increment}.
\end{IEEEproof}

\begin{IEEEproof}[Proof of Theorem~\ref{thm:reference_mean_accuracy}]
The exact RFF representation gives
$\bm{\mu}(\bm{s})=\bm{H}(\bm{s})\bm{\theta}^{\star}$ for an optimal parameter vector
$\bm{\theta}^{\star}\in\mathbb{R}^{Dn_d}$. The residual observations satisfy
\begin{equation}
\bm{d}_k=\bm{H}_k\bm{\theta}^{\star}+\bm{\varepsilon}_k,
\label{eq:reference_mean_proof_observation_model}
\end{equation}
where $\bm{\varepsilon}_k$ denotes
the random deviation from the conditional mean.
Under the contaminated Gaussian model in
\eqref{eq:contaminated_gaussian_residual}, its conditional distribution is
\begin{equation}
\bm{\varepsilon}_k\mid\mathcal{D}_k\sim(1-p_{\zeta_1})\mathcal{N}(\bm{0},\bm{\Sigma}_0)+p_{\zeta_1}\mathcal{N}(\bm{0},\kappa\bm{\Sigma}_0).
\label{eq:reference_mean_proof_noise_model}
\end{equation}
Define
$\bm{\Delta}_k:=\bm{\theta}_k-\bm{\theta}^{\star}$,
$\bm{V}_k:=\bm{P}_k^{-1}$,
$\bar{\bm{H}}_k:=\bm{\Sigma}_0^{-1/2}\bm{H}_k$,
$\bm{z}_k:=\bm{\Sigma}_0^{-1/2}\bm{\varepsilon}_k$,
$\bm{X}_k:=\gamma_k\bar{\bm{H}}_k^\top\bm{z}_k$, and
$\bm{b}_k:=\mathbb{E}[\bm{X}_k\mid\mathcal{D}_k]$.
Assume $\bm{P}_0\succ\bm{0}$,
$\|\bm{\Delta}_0\|_2\le R_0$, and
$\|\bar{\bm{H}}_k\|_2\le L_H<\infty$ for all $k$.
Assume that the features are persistently exciting,
in the sense that $\bm{V}_k\succeq\alpha(k+1)\bm{I}_p$ for some
$\alpha>0$ and all $k\ge0$. Also assume bounded iterates,
$\|\bm{\Delta}_k\|_2\le R<\infty$, and a conditional-bias bound,
$\|\bm{b}_k\|_2\le\beta\|\bm{\Delta}_k\|_2$, for all $k\ge0$,
where $\beta\ge0$ and $c:=\alpha-2\beta>0$.

First, the matrix inversion lemma applied to
\eqref{eq:posterior_weighted_information_update} gives the exact
information recursion
\begin{equation}
\bm{V}_{k+1}
=\bm{V}_k+\gamma_k\bar{\bm{H}}_k^\top\bar{\bm{H}}_k.
\label{eq:reference_mean_information_recursion}
\end{equation}
Consequently, $\bm{V}_{k+1}\succeq\bm{V}_k$ and
$\bm{P}_{k+1}\preceq\bm{P}_k\preceq\bm{P}_0$. The gain in
\eqref{eq:posterior_weighted_rls_gain} can also be written as
$\bm{K}_k=\gamma_k\bm{P}_{k+1}\bm{H}_k^\top\bm{\Sigma}_0^{-1}$,
so that
$\bm{V}_{k+1}\bm{K}_k
=\gamma_k\bm{H}_k^\top\bm{\Sigma}_0^{-1}$.
Since $\bm{e}_k=\bm{\varepsilon}_k-\bm{H}_k\bm{\Delta}_k$,
the parameter update yields
\[
\begin{aligned}
\bm{V}_{k+1}\bm{\Delta}_{k+1}
={}&\bm{V}_{k+1}\bm{\Delta}_k
+\gamma_k\bar{\bm{H}}_k^\top
(\bm{z}_k-\bar{\bm{H}}_k\bm{\Delta}_k)\\
={}&\bm{V}_k\bm{\Delta}_k+\bm{X}_k.
\end{aligned}
\]
Let
$\bm{Z}_k:=\bm{X}_k-\bm{b}_k$ and
$\bm{M}_N:=\sum_{k=0}^{N-1}\bm{Z}_k$.
Then $\mathbb{E}[\bm{Z}_k\mid\mathcal{D}_k]=\bm{0}$, so
$\bm{Z}_k$ is a martingale difference. Summing the preceding identity
gives the error decomposition as
\begin{equation}
\bm{\Delta}_N
=\bm{V}_N^{-1}
\left(
\bm{V}_0\bm{\Delta}_0+
\sum_{k=0}^{N-1}\bm{b}_k+\bm{M}_N
\right).
\label{eq:reference_mean_exact_error_decomposition}
\end{equation}

Next, we obtain a uniform bound on the noise.
Define the whitened innovation and epistemic covariance as
$\bm{r}_k:=\bm{\Sigma}_0^{-1/2}\bm{e}_k
=\bm{z}_k-\bar{\bm{H}}_k\bm{\Delta}_k$ and
$\bar{\bm{A}}_k:=\bar{\bm{H}}_k\bm{P}_k\bar{\bm{H}}_k^\top$,
respectively.
The two whitened innovation likelihoods have covariance matrices
$\bm{I}_{n_d}+\bar{\bm{A}}_k$ and
$\kappa\bm{I}_{n_d}+\bar{\bm{A}}_k$. Taking their likelihood ratio,
as in \eqref{eq:posterior_weight_exponential_form}, gives
\begin{equation}
\gamma_k(\bm{r})
=\frac{1}{1+c_k\exp(\bm{r}^\top\bar{\bm{Q}}_k\bm{r}/2)},
\label{eq:reference_mean_whitened_posterior_weight}
\end{equation}
where
\begin{subequations}
\label{eq:reference_mean_whitened_weight_constants}
\begin{align}
c_k &=\frac{p_{\zeta_1}}{1-p_{\zeta_1}}
\left(
\frac{|\bm{I}_{n_d}+\bar{\bm{A}}_k|}
{|\kappa\bm{I}_{n_d}+\bar{\bm{A}}_k|}
\right)^{1/2},
\label{eq:reference_mean_whitened_weight_c}\\
\bar{\bm{Q}}_k &:=(\bm{I}_{n_d}+\bar{\bm{A}}_k)^{-1}
-(\kappa\bm{I}_{n_d}+\bar{\bm{A}}_k)^{-1}.
\label{eq:reference_mean_whitened_weight_q}
\end{align}
\end{subequations}
Because $\bm{P}_k\preceq\bm{P}_0$ and
$\|\bar{\bm{H}}_k\|_2\le L_H$,
we have $\bm{0}\preceq\bar{\bm{A}}_k\preceq m\bm{I}_{n_d}$,
where $m:=L_H^2\|\bm{P}_0\|_2$.
Thus every eigenvalue $a_i$ of $\bar{\bm{A}}_k$ belongs to $[0,m]$.
The determinant ratio is
$\prod_{i=1}^{n_d}(1+a_i)/(\kappa+a_i)$, and each factor is at
least $1/\kappa$. Also, the corresponding eigenvalue of
$\bar{\bm{Q}}_k$ is
$(\kappa-1)/[(1+a_i)(\kappa+a_i)]$. Therefore,
we have $c_k\ge c_0:=\frac{p_{\zeta_1}}{1-p_{\zeta_1}}\kappa^{-n_d/2}>0$
and $\bar{\bm{Q}}_k\succeq q_0\bm{I}_{n_d}$,
where $q_0:=\frac{\kappa-1}{(1+m)(\kappa+m)}>0$.
It follows that
$\gamma_k(\bm{r})\le\frac{1}{1+c_0\exp(q_0\|\bm{r}\|_2^2/2)}$.
Define
$B_\gamma:=\sup_{t\ge0}\frac{t}{1+c_0\exp(q_0t^2/2)}$.
Using
$\bm{z}_k=\bm{r}_k+\bar{\bm{H}}_k\bm{\Delta}_k$,
$0\le\gamma_k\le1$, and the bounded-iterate condition, we obtain
\begin{equation}
\begin{aligned}
\|\bm{X}_k\|_2
&\le\gamma_k L_H\|\bm{r}_k\|_2
+\gamma_k L_H^2\|\bm{\Delta}_k\|_2\\
&\le L_H(B_\gamma+L_HR).
\end{aligned}
\label{eq:reference_mean_uniform_weighted_noise}
\end{equation}
Define $B_X:=L_H(B_\gamma+L_HR)$.
The bounded-iterate condition controls the displacement between the
innovation and the noise in the learning process.

For each coordinate $j=1,\ldots,p$, the martingale difference
$Z_{k,j}$ has zero conditional mean and, conditional on $\mathcal{D}_k$,
lies in an interval of length $2B_X$.
The conditional-range form of the Azuma--Hoeffding inequality
\cite{hoeffding1963probability} 
therefore gives, for any $\delta\in(0,1)$,
\begin{equation}
\Pr\!\left(|M_{N,j}|>B_X\sqrt{2N}\,\ell_N\right)
\le\frac{\delta}{p(N+1)^2},
\label{eq:reference_mean_coordinate_tail}
\end{equation}
where $\ell_N:=\sqrt{\log\frac{2p(N+1)^2}{\delta}}$.
A union bound over all coordinates and all $N\ge1$ bounds the total
failure probability by
$\sum_{N=1}^{\infty}\sum_{j=1}^{p}\delta/[p(N+1)^2]\le\delta$.
Consequently, with probability at least $1-\delta$, simultaneously for all $N\ge1$,
\begin{equation}
\|\bm{M}_N\|_2\le\sqrt{p}\max_j|M_{N,j}|
\le B_X\sqrt{2pN}\,\ell_N.
\label{eq:reference_mean_uniform_martingale_bound}
\end{equation}

Let $a_0:=\|\bm{V}_0\|_2R_0$.
Whenever \eqref{eq:reference_mean_uniform_martingale_bound} holds for all
$N\ge1$, combining it with \eqref{eq:reference_mean_exact_error_decomposition}
and the assumed information and conditional-bias bounds gives
\begin{equation}
\|\bm{\Delta}_N\|_2
\le
\frac{
a_0+\beta\sum_{k=0}^{N-1}\|\bm{\Delta}_k\|_2
+B_X\sqrt{2pN}\,\ell_N
}{
\alpha(N+1)
}.
\label{eq:reference_mean_error_recursion_bound}
\end{equation}
For $\alpha>2\beta$, set $c:=\alpha-2\beta$ and
$C_N:=(a_0+B_X\sqrt{2p}\,\ell_N)/c$, which is nondecreasing in $N$.
The base case $\|\bm{\Delta}_0\|_2\le C_0$ follows from
$\bm{V}_0\succeq\alpha\bm{I}_p$.
Assuming $\|\bm{\Delta}_k\|_2\le C_k/\sqrt{k+1}$ for every $k<N$ gives
$\sum_{k=0}^{N-1}\|\bm{\Delta}_k\|_2\le2C_N\sqrt{N}$.
Substitution into \eqref{eq:reference_mean_error_recursion_bound},
using $c+2\beta=\alpha$, yields
\begin{equation}
\|\bm{\Delta}_N\|_2
\le\frac{C_N\sqrt{N}}{N+1}
\le\frac{C_N}{\sqrt{N+1}},\qquad N\ge1.
\label{eq:reference_mean_induction_step}
\end{equation}
Thus, induction establishes, with probability at least $1-\delta$
and simultaneously for all $N\ge1$,
\begin{equation}
\|\bm{\Delta}_N\|_2
\le
\frac{
a_0+B_X\sqrt{2p}\,\ell_N
}{
(\alpha-2\beta)\sqrt{N+1}
},
\qquad N\ge1.
\label{eq:reference_mean_parameter_error_bound}
\end{equation}

Finally, let $B_H:=\sup_{\bm{s}\in\mathcal{D}}\|\bm{H}(\bm{s})\|_2$.
For the RFF map in \eqref{eq:rff_map},
\begin{equation}
\|\bm{\phi}(\bm{s})\|_2^2
=\frac{2}{D}\sum_{i=1}^{D}
\cos^2(\bm{\omega}_i^\top\bm{s}+b_i)\le2.
\label{eq:reference_mean_rff_feature_norm_bound}
\end{equation}
Since $\bm{H}(\bm{s})
=\bm{I}_{n_d}\otimes\bm{\phi}(\bm{s})^\top$,
we have $B_H\le\sqrt{2}$.
Thus, with probability at least $1-\delta$, simultaneously for all
$N\ge1$,
\begin{equation}
\sup_{\bm{s}\in\mathcal{D}}
\|\widehat{\bm{\mu}}_N(\bm{s})-\bm{\mu}(\bm{s})\|_2
\le
\frac{B_H\left(a_0+B_X\sqrt{2p}\,\ell_N\right)}
{(\alpha-2\beta)\sqrt{N+1}}.
\label{eq:reference_mean_explicit_prediction_bound}
\end{equation}
For fixed $p$, $\delta$, and the remaining constants, the right-hand
side of \eqref{eq:reference_mean_explicit_prediction_bound} is of order
$\mathcal{O}\!\left(\sqrt{\frac{\log(N+1)}{N+1}}\right)$.
\end{IEEEproof}

\begin{IEEEproof}[Proof of Lemma~\ref{lem:residual_compensation_error}]
Evaluating the parameter error bound
\eqref{eq:reference_mean_parameter_error_bound} at $N=k$ gives
\[
\|\bm\theta^\star-\bm\theta_k\|_2
\le \frac{C_k}{\sqrt{k+1}},
\]
where $C_k$ is the coefficient in
\eqref{eq:reference_mean_induction_step}.
Choose the uncertainty scale as
\begin{equation}
\beta_k:=
\sqrt{\lambda_{\max}(\bm P_k^{-1})}\,
\frac{C_k}{\sqrt{k+1}}.
\label{eq:rollout_confidence_calibration}
\end{equation}
Then it follows that
\[
\|\bm P_k^{-1/2}
(\bm\theta^\star-\bm\theta_k)\|_2
\le \beta_k.
\]

For any $\bm s_{i\mid k}\in\mathcal D$, let
$\bm H_{i\mid k}:=\bm H(\bm s_{i\mid k})$.
Using the RFF representation and
$\bm A_{i\mid k}=\bm H_{i\mid k}\bm P_k\bm H_{i\mid k}^{\top}$,
we obtain
\begin{equation}
\begin{aligned}
&\|\bm\mu(\bm s_{i\mid k})-\hat{\bm\mu}_{i\mid k}
\|_{\bm\Sigma_0^{-1}}\\
&\quad=
\|\bm\Sigma_0^{-1/2}\bm H_{i\mid k}
\bm P_k^{1/2}\bm P_k^{-1/2}
(\bm\theta^\star-\bm\theta_k)\|_2\\
&\quad\le
\beta_k\|\bm\Sigma_0^{-1/2}
\bm H_{i\mid k}\bm P_k^{1/2}\|_2\\
&\quad=
\beta_k\sqrt{\lambda_{\max}\!\left(
\bm\Sigma_0^{-1/2}\bm A_{i\mid k}
\bm\Sigma_0^{-1/2}\right)}\\
&\quad=r_{i\mid k},
\end{aligned}
\end{equation}
where $r_{i\mid k}$ is the calibrated confidence radius
in \eqref{eq:mpc_residual_uncertainty_radius}.

The Cauchy--Schwarz inequality
and the confidence bound above give
\[
\hat{\bm\mu}_{i\mid k}^{\top}
\bm\Sigma_0^{-1}\bm\mu(\bm s_{i\mid k})
\ge m_{i\mid k}^2-m_{i\mid k}r_{i\mid k}
=m_{i\mid k}(m_{i\mid k}-r_{i\mid k}).
\]
Using $\eta_{i\mid k}$ defined in
\eqref{eq:uncertainty_adaptive_residual_gain}, we obtain
\[
\begin{aligned}
&\|\bm e^{\mathrm{comp}}_{i\mid k}
\|_{\bm\Sigma_0^{-1}}^2\\
&\quad=
\|\bm\mu(\bm s_{i\mid k})\|_{\bm\Sigma_0^{-1}}^2
+\eta_{i\mid k}^2m_{i\mid k}^2
-2\eta_{i\mid k}\hat{\bm\mu}_{i\mid k}^{\top}
\bm\Sigma_0^{-1}\bm\mu(\bm s_{i\mid k})\\
&\quad\le
\|\bm\mu(\bm s_{i\mid k})\|_{\bm\Sigma_0^{-1}}^2
+\eta_{i\mid k}^2m_{i\mid k}^2
-2\eta_{i\mid k}m_{i\mid k}
(m_{i\mid k}-r_{i\mid k})\\
&\quad=
\|\bm\mu(\bm s_{i\mid k})\|_{\bm\Sigma_0^{-1}}^2
-(m_{i\mid k}-r_{i\mid k})^2.
\end{aligned}
\]
\end{IEEEproof}

\vfill

\end{document}